%% file: main.tex
\documentclass{article}

\usepackage{xr-hyper}
\usepackage{hyperref}

\usepackage[preprint]{arxiv}

\usepackage[utf8]{inputenc} 
\usepackage[T1]{fontenc}    
\usepackage{url}            
\usepackage{booktabs}       
\usepackage{amsfonts}       
\usepackage{nicefrac}       
\usepackage{microtype}      
\usepackage{xcolor}         
\usepackage{bm}
\usepackage{wrapfig}
\usepackage{comment}

\usepackage[bottom]{footmisc}
\usepackage{xspace}
\usepackage{tikz} 
\usepackage{dsfont} 
\usetikzlibrary{bayesnet,calc,decorations.pathreplacing}
\usepackage{amsmath,amssymb}
\usepackage{subcaption}
\usepackage[numbers]{natbib}
\usepackage[nolist,smaller]{acronym}
\input{macros_common}
\input{macros_math}

\input{control_flow_variables}
\usepackage{intcalc}
\usepackage{mathtools}
\usepackage{cancel}
\usepackage{ifthen}
\usepackage{listings}
\usepackage[frozencache=true,cachedir=minted-cache]{minted} 
\usepackage{amsthm}
\newtheorem{proposition}{Proposition}
\newtheorem{innernumberedproposition}{Proposition}
\newenvironment{numberedproposition}[1]
  {\renewcommand\theinnernumberedproposition{#1}\innernumberedproposition}
  {\endinnernumberedproposition}

\ifapendixinsamepdf
{}
\else
{
\makeatletter
\newcommand*{\addFileDependency}[1]{
  \typeout{(#1)}
  \@addtofilelist{#1}
  \IfFileExists{#1}{}{\typeout{No file #1.}}
}
\makeatother

\newcommand*{\myexternaldocument}[1]{%
    \externaldocument{#1}%
    \addFileDependency{#1.tex}%
    \addFileDependency{#1.aux}%
}
\myexternaldocument{appendix}
}
\fi

\title{Efficient Transformed Gaussian Processes for Non-Stationary Dependent Multi-class Classification}

\author{%
  Juan Maroñas \\
  Machine Learning Group\\
  Universidad Autónoma de Madrid\\
  Madrid, Spain \\
  \texttt{juan.maronnas@uam.es} \\
  \And
  Daniel Hernández-Lobato \\
  Machine Learning Group\\
  Universidad Autónoma de Madrid\\
  Madrid, Spain \\
 \texttt{daniel.hernandez@uam.es} \\
}

\begin{document}
\normalsize

\maketitle

\begin{abstract}
\looseness=-1 This work introduces the Efficient Transformed Gaussian Process 
(\ETGP), a new way of creating $C$ stochastic processes characterized by:
1) the $C$ processes are \emph{non-stationary}, 
2) the $C$ processes are dependent by construction without needing a mixing matrix,
3) training and making predictions is very efficient since the number of 
Gaussian Processes (\GP) operations (\eg inverting the inducing point's covariance 
matrix) do not depend on the number of processes.  This makes the 
\ETGP particularly suited for multi-class problems with a very 
large number of classes, which are the problems studied in this work.
\ETGP{}s exploit the recently proposed Transformed Gaussian Process (\TGP), a 
stochastic process specified by transforming a Gaussian Process using an 
invertible transformation. However, unlike \TGP{}s, \ETGP{}s are constructed 
by transforming a single sample from a \GP using $C$ invertible transformations. 
We derive an efficient sparse variational inference algorithm for the proposed 
model and demonstrate its utility in 5 classification tasks which include 
low/medium/large datasets and a different number of classes, ranging from 
just a few to hundreds. Our results show that \ETGP{}s, in general, outperform 
state-of-the-art methods for multi-class classification 
based on \GP{}s, and have a lower computational cost 
(around one order of magnitude smaller).
\end{abstract}

\input{sections/A_introduction}
\input{sections/B_background}
\input{sections/C_Model_specification}

\input{sections/D_inference}
\input{sections/E_related_work}

\input{sections/F_experiments}

\input{sections/G_conclusions_future_work}

\ifsocietalimpactinmain
  \input{sections/Z011_societal_impact}
\fi 

\ifincludeacks
    \input{sections/Z01_acks}
\fi


\input{main.bbl}
\ifethicsquestionsinpdf
{
\input{sections/Z0_ethics}

\clearpage
}
\else
{}
\fi

\ifapendixinsamepdf
{

\clearpage

\title{Appendix for Efficient Transformed Gaussian Processes for Non-Stationary Dependent Multi-class Classification}

\maketitleapx

\appendix

\input{sections/Z1_equations_appendix}
\clearpage
\input{sections/Z2_experiments_appendix}
\clearpage
\input{sections/Z3_refactorizing_gpflow_source_code}

}
\fi

\end{document}

%% file: macros_common.tex
\definecolor{darkpink}{rgb}{0.91, 0.33, 0.5}
\definecolor{darkgreen}{rgb}{0.0, 0.5, 0.0}

\newcommand{\usec}{Sec.\xspace}
\newcommand{\usubsec}{Sec.\xspace}

\newcommand{\ueqn}{Eq.\xspace}
\newcommand{\uappendix}{App.\xspace}
\newcommand{\fig}{Fig.\xspace}
\newcommand{\prop}{Prop.\xspace}
\newcommand{\acro}[1]{\textsc{#1}\xspace}

\newcommand{\GP}{\acro{\smaller GP}}
\newcommand{\ETGP}{\acro{\smaller ETGP}}
\newcommand{\TGP}{\acro{\smaller TGP}}
\newcommand{\DGP}{\acro{\smaller DGP}}
\newcommand{\NN}{\acro{\smaller NN}}

\newcommand{\SOTA}{\acro{\smaller SOTA}}
\newcommand{\GPFLOW}{\acro{\smaller GPFLOW}}
\newcommand{\UCI}{\acro{\smaller UCI}}
\newcommand{\BNN}{\acro{\smaller BNN}}
\newcommand{\SVI}{\acro{\smaller SVI}}
\newcommand{\VI}{\acro{\smaller VI}}
\newcommand{\ELBO}{\acro{\smaller ELBO}}
\newcommand{\KLD}{\acro{\smaller KLD}}
\newcommand{\ELL}{\acro{\smaller ELL}}
\newcommand{\SVGP}{\acro{\smaller SVGP}}
\newcommand{\DNN}{\acro{\smaller DNN}}
\newcommand{\MCDROP}{\acro{\smaller MCDROP}}
\newcommand{\ACC}{\acro{\smaller ACC}}
\newcommand{\LL}{\acro{\smaller LL}}
\newcommand{\LINEAR}{\acro{\smaller LINEAR}}
\newcommand{\SAL}{\acro{\smaller SAL}}
\newcommand{\TANH}{\acro{\smaller TANH}}
\newcommand{\LOTUS}{\acro{\smaller LOTUS}}
\newcommand{\GPYTORCH}{\acro{\smaller GPYTORCH}}
\newcommand{\TENSORFLOW}{\acro{\smaller TENSORFLOW}}
\newcommand{\LMC}{\acro{\smaller LMC}}

\newcommand{\BAETGP}{\acro{\smaller BA-ETGP}}
\newcommand{\PEETGP}{\acro{\smaller PE-ETGP}}

\newcommand{\RBF}{\acro{\smaller RBF}}
\newcommand{\RBFCORR}{\acro{\smaller RBFCORR}}
\newcommand{\ARCCOS}{\acro{\smaller ARCCOS}}
\newcommand{\NNET}{\acro{\smaller NNET}}

\newcommand{\aka}{a.k.a.\xspace}
\newcommand{\etc}{etc\xspace}
\newcommand{\wrt}{w.r.t.\xspace}
\newcommand{\eg}{e.g.\xspace}
\newcommand{\ie}{i.e.\xspace}

%% file: macros_math.tex
\newcommand{\mb}[1]{\mathbf{#1}}

\newcommand{\Expectation}[2]{\mathds{E}_{#1}\left[#2\right]}

\newcommand{\complexity}[1]{\mathcal{O}(#1)}

\newcommand{\Ngauss}[1]{\mathcal{N}\left(#1\right)}
\newcommand{\Ngaussshort}[1]{\mathcal{N}(#1)}

\newcommand{\Xspace}{\mathcal{X}\xspace} 
\newcommand{\X}[1]{\mb{x}^{#1}\xspace} 
\newcommand{\Xsamples}{\mb{X}\xspace} 

\newcommand{\Z}[1]{\mb{z}^{#1}\xspace} 
\newcommand{\Zsamples}[1]{\mb{Z}^{#1}\xspace}

\newcommand{\Zallsamples}[1]{\overline{\mb{Z}}^{#1}\xspace} 

\newcommand{\Yspace}{\mathcal{Y}\xspace} 
\newcommand{\Y}[1]{y^{#1}\xspace} 
\newcommand{\Ysamples}{\mb{y}\xspace} 

\newcommand{\Dsamples}{\mathcal{D}\xspace} 

\newcommand{\veczero}{\mb{0}}
\newcommand{\Sinv}{\mb{S}^{-1}}
\newcommand{\Q}{\mb{Q}\xspace}
\newcommand{\Qinv}{\mb{Q}^{-1}\xspace}
\newcommand{\rrr}{\mb{r}\xspace}

\newcommand{\varm}[1]{\mathbf{m}^{#1}}
\newcommand{\varS}[1]{\mathbf{S}^{#1}}

\newcommand{\varmall}{\overline{\mathbf{m}}}
\newcommand{\varSall}{\overline{\mathbf{S}}}

\newcommand{\myprod}[2]{\prod_{#1}^{#2}} 
\newcommand{\mysum}[2]{\sum_{#1}^{#2}} 

\newcommand{\myprodinline}[2]{\textstyle \prod_{#1}^{#2}} 
\newcommand{\mysuminline}[2]{\textstyle \sum_{#1}^{#2}} 

\newcommand{\dd}[1]{\text{d}#1}

\newcommand{\A}[1]{\mb{A}_{#1}\xspace}
\newcommand{\Ainv}[1]{\mb{A}^{-1}_{#1}\xspace}
\newcommand{\aaa}[1]{\mb{a}^{#1}\xspace}
\newcommand{\bb}[1]{\mb{b}^{#1}\xspace}

\newcommand{\aanb}[2]{a_{#1}^{#2}\xspace}
\newcommand{\bbnb}[2]{b_{#1}^{#2}\xspace}

\newcommand{\GPcovar}[1]{K^{#1}_\nu}
\newcommand{\nuc}[1]{\nu^{#1}}
\newcommand{\nuall}{\overline{\nu}}

\newcommand{\lambdac}[1]{\lambda^{#1}}
\newcommand{\lambdaall}{\overline{\lambda}}
\newcommand{\phiall}{\overline{\phi}}

\newcommand{\W}[1]{\mb{W}^{#1}\xspace}
\newcommand{\Wall}{\mb{\overline{W}}\xspace}

\newcommand{\f}[1]{\mb{f}^{#1}\xspace} 

\newcommand{\fK}[1]{\mb{f}^{#1}_K\xspace} 
\newcommand{\fnull}[1]{\mb{f}^{#1}_0\xspace} 
\newcommand{\fpos}[2]{\mb{f}^{#2}_{#1}\xspace} 

\newcommand{\fall}{\mb{\overline{f}}\xspace} 
\newcommand{\fallk}{\mb{\overline{f}}_K\xspace} 
\newcommand{\fallnull}{\mb{\overline{f}}_0\xspace} 

\newcommand{\fnb}[1]{f^{#1}\xspace} 

\newcommand{\fnbK}[1]{f^{#1}_K\xspace}
\newcommand{\fnbnull}[1]{f^{#1}_0\xspace} 
\newcommand{\fnbpos}[2]{f^{#2}_{#1}\xspace}

\newcommand{\fnballpos}[2]{\overline{f}^{#2}_{#1}\xspace}

\newcommand{\uu}[1]{\mb{u}^{#1}\xspace} 

\newcommand{\uK}[1]{\mb{u}^{#1}_K\xspace} 
\newcommand{\unull}[1]{\mb{u}^{#1}_0\xspace} 
\newcommand{\upos}[2]{\mb{u}^{#2}_{#1}\xspace} 

\newcommand{\uall}{\mb{\overline{u}}\xspace} 
\newcommand{\uallk}{\mb{\overline{u}}_K\xspace} 
\newcommand{\uallnull}{\mb{\overline{u}}_0\xspace} 

\newcommand{\unbnull}[1]{u^{#1}_0\xspace}

\newcommand{\G}{\mathbb{G}}
\newcommand{\Gk}[1]{\mathbb{G}_{#1}}
\newcommand{\Gcomp}{\Gk{\theta_K}}

\newcommand{\Gc}[1]{\mathbb{G}^{#1}}

\newcommand{\Gcompc}[1]{\Gk{\theta_K}^{#1}}
\newcommand{\Gcompcinpdep}[2]{\Gk{\theta_K(\W{#1},#2)}^{#1}}

\newcommand{\Gall}{\overline{\mathbb{G}}}
\newcommand{\Gallcomp}{\overline{\mathbb{G}}_{\theta_K}}
\newcommand{\Gallcompinpdep}[1]{\overline{\mathbb{G}}_{\theta_K(\Wall,#1)}}

\newcommand{\HH}{\mathbb{H}}
\newcommand{\Hk}[1]{\mathbb{H}_{#1}}

\newcommand{\Hcompc}[1]{\Hk{\theta_K}^{#1}}

\newcommand{\Hcompcinpdep}[2]{\Hk{\theta_K(\W{#1},#2)}^{#1}}

\newcommand{\Gcdirinvcomp}[2]{\Gcompc{#1} \circ \Hcompc{#2}}
\newcommand{\Gcdirinvcompinpdep}[3]{\Gcompcinpdep{#1}{#3} \circ \Hcompcinpdep{#2}{#3}}

\newcommand{\Jacobiancinputdep}[3]{\myprod{k=0}{K^{#2}-1}\left| \det \frac{\partial \Gc{#2}_{\theta_k(\W{#2},#3)}(#1)}{\partial #1} \right|^{-1}}

\newcommand{\Jacobiancinputdepshortinline}[3]{\myprodinline{k=0}{K^{#2}-1}| \det \frac{\partial \Gc{#2}_{\theta_k(\W{#2},#3)}(#1)}{\partial #1} |^{-1}}

\newcommand{\mydeltafinpdep}[3]{\delta\left(\fK{#1} - \Gcdirinvcompinpdep{#1}{#2}{#3}(\fK{#2})\right)}
\newcommand{\mydeltafinpdepshort}[3]{\delta(\fK{#1} - \Gcdirinvcompinpdep{#1}{#2}{#3}(\fK{#2}))}

\newcommand{\mydeltau}[2]{\delta\left(\uK{#1} - \Gcdirinvcomp{#1}{#2}(\uK{#2})\right)}

\newcommand{\mydeltauinpdep}[3]{\delta\left(\uK{#1} - \Gcdirinvcompinpdep{#1}{#2}{#3}(\uK{#2})\right)}
\newcommand{\mydeltauinpdepshort}[3]{\delta(\uK{#1} - \Gcdirinvcompinpdep{#1}{#2}{#3}(\uK{#2}))}



%% file: control_flow_variables.tex
\newif\ifapendixsubmission 
\newif\ifapendixinsamepdf 
\newif\ifethicsquestionsinpdf 

\newif\ifsocietalimpactinmain 
\newif\ifsocietalimpactinapx  

\newif\ifincludeacks  

\apendixsubmissionfalse 

\apendixinsamepdftrue
\ethicsquestionsinpdffalse

\societalimpactinapxfalse
\societalimpactinmainfalse

\includeackstrue

%% file: sections/A_introduction.tex
\section{Introduction}

\looseness=-1 Gaussian Processes (\GP{}s) are stochastic processes 
characterized by their finite-dimensional distributions being multivariate 
Gaussian \citep{GPbook}, and have become a uniquely popular 
modeling tool. For example, in the machine learning 
community \GP{}s are used as  prior distributions over 
functions, used to solve tasks 
such as regression, classification, feature extraction or 
hyper-parameter optimization \citep{GPbook,lawrence2003gaussian,snoek2012practical}.
Their non-parametric nature imply
that they become more expressive
with more data \citep{GPbook}. Furthermore, \GP{}s are characterized by a predictive
distribution which provides information about what the model does not know \citep{Gal2016Uncertainty} 
and are
easy to interpret since the covariance function 
gives insights about the nature of the latent function to be inferred \citep{duvenaud2013structure}. 
\GP{}s
have also been applied in spatial statistics \citep{kriging}, and to 
explain physics phenomena such as those that arise when studying molecular dynamics 
\citep{book_molecular_dynamics,OUProcess}. Moreover, they are used as a theoretical 
tool to understand Deep Neural Networks (\DNN) \citep{NealPhd,NNareGpsGreg} and lie 
at the core of a recent family of Deep Generative Models that generate samples 
attending to the dynamics of a diffusion process \citep{song2021scorebased}.

Here, we focus on multi-class classification problems with $C>2$ classes. For this, one 
often defines $C$ independent \GP{}s, one per each class \citep{GPbook}. 
In this case, the number of \GP{} operations (such as inverting the kernel matrix over the inducing points) 
grows linearly with $C$. Thus, if $C$ is large, this can be too expensive. 
Some speed-up tricks include sharing the inducing points or the kernel across each \GP{}
but often reduce the performance of the classifier, as we show in our experiments. 
Even with this trick, computing the parameters of the 
predictive distribution still has complexity $\complexity{CM^2}$ per datapoint.
We can gain additional performance by defining
a prior using $C$ dependent \GP{}s, 
which can be done by mixing $Q$ latent \GP{}s with a mixing matrix 
$\bm \Phi \in \mathds{R}^{Q\times C}$. However, in practice, these dependencies 
are often ignored since the memory complexity scales as $\complexity{C^2}$ per datapoint. 
In fact, modern \SOTA \GP{} software's like \GPFLOW \citep{GPflow2017,GPflow2020multioutput} 
require significant source code modification (up to early 2022) to handle these dependencies 
in an efficient way. 

A disadvantage of \GP{}s is that they usually impose strong assumptions about the
nature of the latent function. For example, most covariance functions are stationary and assume a constant level of smoothness for the latent function on the input domain \citep{GPbook}. If this is not the case, the performance can be degraded. However, \GP{}s can be made more expressive using non-stationary processes, but this is usually only justified if one has background knowledge about the nonstationarity of the particular application. For example, for Bayesian Optimization \citep{InputWarpingBO}, Geostatistics \citep{KangruiNSNS,GPRNwilson,OllieMRMT,sampson1992nonparametric,Schmidt00bayesianinference} or temporal gene expression \citep{NScovheinonen16}.

The flexibility of \GP{}s can also be increased by non-linearly transforming these processes. Examples include deep \GP{}s (\DGP{}s) \citep{DGP_seminal} and transformed \GP{}s
(\TGP{}s) \citep{TGP_maronas}. In \DGP{}s, the output of a \GP{} is used as the input of
another \GP systematically, following a fully connected neural network (\NN{}) architecture in which units are \GP{}s. As a result of the concatenation, the resulting process need not be
stationary. In \TGP{}s the initial GP prior is transformed iteratively using input-dependent invertible 
transformations \citep{TGP_maronas}. Because of this input dependence, the resulting process
need not be stationary. Importantly, \TGP{}s often generate models that are 
as accurate as \DGP{}s at a lower computational cost. 

In this work we introduce the Efficient Transformed Gaussian Process (\ETGP), a 
new model where $C$ processes are specified by sampling from a single \GP{}, and then 
transforming this sample using $C$ invertible transformations (throughout the 
paper we refer to the invertible transformations by flows or warping functions as well). 
By this construction, the $C$ processes are non-stationary and dependent, with dependencies 
modeled by the copula of the base \GP{} and without the computational and memory complexity of an equivalent number of \GP{}s, since only one $\GP{}$ is used in the construction of the $C$ processes. 
A special case of the \ETGP family specified by using a linear flow includes non-stationary 
dependent \GP{}s, which we also characterize and study.  We evaluate the prediction 
performance and computational cost of \ETGP in the context of multi-class problems with a 
large number of classes $C$. With this goal, we derive an efficient sparse variational inference (\VI) 
algorithm for \ETGP. We carry out experiments across several classification tasks which
include small and large datasets and up to 153 class labels. The results obtained show 
that \ETGP{}s, in general, outperform \SOTA methods for multi-class 
classification based on \GP{}s, and that they have a computational cost 
that is around one order of magnitude smaller. Our experiments also show  
that non-stationary covariances are not useful for black-box function 
approximations and that the particular inductive bias of the \ETGP is rather 
much more beneficial.

%% file: sections/B_background.tex
\section{Background}
\label{sec:back}

We start by introducing \GP{}s for multi-class classification problems 
and some notation. We also describe how to improve \GP{}s using the
\TGP method.

\subsection{Multi-class Gaussian process classification}
\label{sec:background_mgp}
Consider the problem of assigning a class label $\Y{} \in \Yspace = \{1,\ldots,C\}$, with $C$ the number 
of classes, to an input $\X{} \in \Xspace  \subseteq \mathds{R}^{d}$. Our goal is to learn a set of $C$ functions mapping 
 $\X{}$ to class label probabilities. For this, we are given a set of $N$ labeled instances
$\Dsamples=\left\{\X{n},\Y{n}\right\}^N_{n=1}$ 
generated from the data distribution, and define $\Xsamples=(\X{1},\ldots,\X{N})$ and $\Ysamples=(\Y{1},\ldots,\Y{N})$. We model these functions in a Bayesian way by placing an independent \GP{} over each of them, which is updated into a posterior over functions given $\Dsamples$, used to obtain a predictive distribution for the label associated to new data. 

A \GP{} is a stochastic process whose finite-dimensional distributions are 
given by a multivariate Gaussian. Specifically, let $\f{}=(\fnb{}(\X{1}),\ldots,\fnb{}(\X{N}))^\text{T}$ and define $\fnbpos{n}{} \coloneqq \fnb{}(\X{n})$.  Then, $\f{} \sim \Ngaussshort{\mu_\nu(\Xsamples{}),\GPcovar{}(\Xsamples{},\Xsamples{})}$, where the mean vector $\mu_\nu(\Xsamples{})=(\mu_\nu(\X{1}),\ldots,\mu_\nu(\X{N}))^\text{T}$ is obtained by a mean function $\mu_\nu: \Xspace \rightarrow \mathds{R}$, and $\GPcovar{}(\X{},\X{})$ is a $N \times N$ matrix whose $i$-th row and $j$-th column are given by
$\GPcovar{}(\X{i},\X{j})$, obtained by a covariance function  $K_\nu: \Xspace \times \Xspace \rightarrow \mathds{R}$; both parameterized by $\nu$. With out loss of 
generality, in the rest of the paper we assume zero mean \GP{}s.

Consider $C$ independent \GP{}s and denote $\fall{}=\{\f{1},\hdots\f{C}\}$, where a bar over a letter $\overline{x}$ summarizes the corresponding $C$ elements $x^1,\hdots, x^C$. Often, a Softmax link function
$\pi_c(\fnballpos{n}{}) = \exp(\fnb{c}(\X{n}) / \mysuminline{c'=1}{C} \exp(\fnb{c'}(\X{n}))$
is applied to $\fall{}$ to turn 
them into class label probabilities $\pi_c$ \citep{GPbook}. Then, these probabilities 
are linked to actual class labels $\Ysamples{}$ by a categorical likelihood. 
Under these conditions, the joint distribution of $\Ysamples{}$ and $\fall{}$ is:
\begin{align}
    p(\Ysamples{},\fall{}) & = 
	p(\Ysamples{}\mid\fall{}) p(\fall{}) = 
	\left[ 
	\myprodinline{n=1}{N} \myprodinline{c=1}{C} \pi_c(\fnballpos{n}{})^{\mathbb{I}(\Y{n}=c)} \right]
	\myprodinline{c=1}{C}\Ngaussshort{\f{c}\mid\veczero,\GPcovar{c}(\Xsamples,\Xsamples)},
\end{align}
where $\mathbb{I}$ represents the indicator function. To make predictions we need to approximate the intractable posterior $p(\fall{}\mid\Dsamples)$. We rely on the sparse variational inference (\VI) approximation \citep{chai2012variational,VIinducingpoints_titsias} with the modifications introduced 
in \citep{GPsBigData_hensman} to scale to very large datasets. 

Sparse variational \GP{}s (\SVGP{}s) work by introducing a set of $M\ll N$ inducing points locations $\Zsamples{}=(\Z{1},\ldots,\Z{M}), \Z{} \in \Xspace$ with associated \GP outputs
$\uu{}=(f(\Z{1}),\ldots,f(\Z{M}))^\text{T}$ per each of the $C$ functions, with joint Gaussian prior $p(\fall{},\uall{}\mid\Xsamples{},\Zallsamples{})$ obtained with the prior covariance function.
These inducing points act as sufficient statistics of the data $\X{}$, with the purpose of representing the posterior $p(\fall{}\mid\Dsamples)$ using $M$ points, reducing the complexity from $\complexity{CN^3}$ \citep{GPbook} to $\complexity{CM^3}$ \citep{VIinducingpoints_titsias}. The key point in \citep{VIinducingpoints_titsias} is to treat $\Zallsamples{}{}$ as variational parameters, which are optimized by minimizing the Kullback-Leibler Divergence (\KLD) between a variational posterior $q(\fall{},\uall{})$ and the augmented joint posterior $p(\fall,\uall{}\mid\Dsamples,\Zallsamples{})$, or equivalently by maximizing the Evidence Lower Bound (\ELBO). Since $\Zallsamples{}$ are variational parameters, they are protected from overfitting. The 
speed-up is achieved by constraining the form of the
variational distribution $q(\fall{},\uall{})=\myprod{c=1}{C} 
p(\f{c}\mid\uu{c})q(\uu{c})$, which  is 
defined using the conditional model's prior $p(\f{c}\mid\uu{c})$ and a Gaussian variational distribution $q(\uu{c})$ with mean and covariance matrix $\varm{c} \in \mathds{R}^M$ and 
$\varS{c} \in \mathds{R}^{M\times M}$. 
With this, the \ELBO is:
\begin{align}
    \ELBO & = \mysuminline{n=1}{N}\mysuminline{c=1}{C} \mathbb{I}(\Y{n}=c) \mathds{E}_q\left[ \log \pi_c(\fnballpos{n}{})\right] 
	- \mysuminline{c=1}{C}\KLD[q(\uu{c})\mid\mid p(\uu{c}))]\,, 
	\label{elbo_multiclass_independent_gp}
\end{align}
where $\KLD[q(\uu{c})\mid\mid p(\uu{c}))]$ can 
be computed in closed form and the expectation with respect to $q$ can be approximated by Monte Carlo. The above expression allows to use stochastic \VI to optimize the \ELBO \citep{GPsBigData_hensman}, by sub-sampling the data using mini-batches \citep{GPsBigData_hensman}. We use path-wise derivatives for black-box low-variance gradient estimations.
The variational distribution 
$q(\fnb{c}(\X{n})) = \int p(\fnb{c}(\X{n})\mid\mathbf{u}^{c})q(\uu{c})\dd \uu{c}$ is Gaussian 
with mean and covariance given by $\GPcovar{c}(\X{n},\Zsamples{c}) \GPcovar{c}({\Zsamples{c},\Zsamples{c}})^{-1}\varm{c}$ 
and $\GPcovar{c}(\X{n},\X{n})-\GPcovar{c}({\X{n},\Zsamples{c}})\GPcovar{c}({\Zsamples{c},\Zsamples{c}})^{-1}[\GPcovar{c}({\Zsamples{c},\Zsamples{c}})+\varS{c}]\GPcovar{c}({\Zsamples{c},\Zsamples{c}})^{-1}\GPcovar{c}(\Zsamples{c},\X{n})$, respectively. The predictive distribution for the label $\Y{\star}$ associated to a new point $\X{\star}$, can also be approximated via Monte Carlo 
by sampling from $q(\fnb{c}(\X{\star}))$ for $c=1,\ldots,C$. 


\subsection{Transformed Gaussian Processes}
\label{sec:background_tgp}
\looseness=-1 A limitation of \GP{}s is that they place strong assumptions over the latent function. This can be 
for example assuming the same level of smoothness across the input domain, as it is often done by 
considering a stationary covariance function \citep{GPbook}. The Transformed Gaussian Process (\TGP) \citep{TGP_maronas}
is a model that increases the flexibility of \GP{}s by non-linearly and input-dependently transforming 
the \GP prior using invertible transformations. We describe the \TGP, since is the building block of the proposed approach.

Let $\fnbnull{}(\cdot) \sim \GP(0,K_\nu(\cdot,\cdot))$ be a sample from a \GP. Consider a composition 
of individual invertible functional mappings  
$\Gk{\theta_K} = \Gk{\theta_0}\circ \Gk{\theta_1} \hdots \circ \Gk{\theta_{K-1}} : \mathcal{F}_0 \rightarrow \mathcal{F}_K$ 
each parameterized by $\theta \in \Theta$. The \TGP,
is defined by the following generative procedure:
\begin{align}
    \fnbnull{}(\cdot) &\sim \GP(0,K_\nu(\cdot,\cdot))\,, &
    \fnbK{}(\cdot) &= \Gcomp{}(\fnbnull{}(\cdot))\,.
\end{align}
An easy way to specify $ \Gcomp{}$ so that $f_K$ is a consistent process 
(\ie a process satisfying the Kolmogorov extension theorem) is to use 
element-wise mappings (also known by diagonal flows), characterized 
by: $\forall\, \X{n} \in \Xspace$, $\fnbK{}(\X{n})=
\Gcomp{(\fnbnull{}(\X{n}))}$, as described in \citep{riosTransportGPs}\footnote{The 
presented method and inference algorithm, and all the equations involved, assumes diagonal flows. One needs 
to check particularities that might arise for non-diagonal ones, see \eg the appendix of \citep{TGP_maronas}. 
Furthermore, any additional details about the equations in this paper and their derivation are given 
in \uappendix \ref{math_appendix}.}. Due to these element-wise mappings, the resulting process is 
a Gaussian Copula process \citep{CPWilson} since it has arbitrary marginals with dependencies driven by 
the copula of the \GP, something derived from Sklar's theorem \citep{SklarTheorem}. In other words, 
$\fnbnull{}$ and $\fnbK{}$ share the same dependencies but differ in their marginal distributions.
By increasing $K$, we can make the flow as complicated as we want, increasing flexibility.

\begin{figure}[!t]
\begin{subfigure}[b]{.49\textwidth}
\input{images/PGM_TGP}
\end{subfigure}
\begin{subfigure}[b]{.49\textwidth}
\input{images/PGM_ETGP}    
\end{subfigure}
\caption{Probabilistic Graphical models for the \TGP (left) and the \ETGP (right). }
\label{ETGPTGPPGM}
\end{figure}
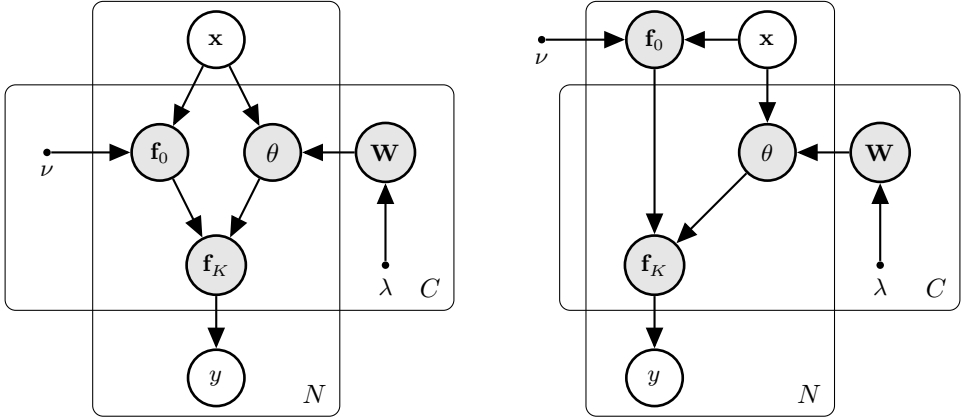

The work in \citep{TGP_maronas} shows how to create non-stationary processes 
$\fnbK{}$ by making the parameters of the flow depend on the input using
a Neural Network (\NN) $\NN : \Xspace \rightarrow \Theta$ parameterized by 
$\W{}$. In \citep{TGP_maronas} they show that this non-stationary process is way more expressive 
than a \GP{} and also than a stationary \TGP{}. Since the parameters of the \NN play the same 
role of hyper-parameters in a \GP{}, this work shows how using a Bayesian Neural 
Network (\BNN) 
 can effectively avoid over-fitting as this makes the \NN's parameters 
play the same role in the graphical model as \GP{}'s latent functions, see \fig \ref{ETGPTGPPGM}.

For a classification problem, we can write the joint conditional distribution over the $C$ independent \TGP{}s by applying the change of variable formula and inverse function theorem \citep{TGP_maronas}:
\begin{align}
\begin{split}
 p(\fallk\mid\Wall{})=\myprodinline{c=1}{C} p(\fnull{}\mid\nuc{c})\Jacobiancinputdepshortinline{\fpos{k}{c}}{c}{\Xsamples{}}.
 \label{etgp_prior}
\end{split}
\end{align}
This warping procedure  can also be used when defining a variational approximation 
$q(\fK{},\uK{})=p(\fK{}\mid\uK{})q(\uK{})$,
where the cancellations of several factors in the \ELBO result in an efficient training
algorithm, as described in \citep{TGP_maronas}.

%% file: images/PGM_TGP.tex
    \begin{tikzpicture}[baseline]
    
    \tikzstyle{RV}=[shape=circle,line width=1.0pt,minimum size = 0.75cm, draw=black,fill=gray!20]
    \tikzstyle{RVo}=[shape=circle,line width=1.0pt,minimum size = 0.75cm, draw=black,fill=white]
    \tikzstyle{param}=[fill,circle,inner sep=1.0pt,minimum size=0.5pt]
    \tikzstyle{plate}=[draw, rectangle, rounded corners]

    \node [RVo] (X) at (2.25,1.5) {\small$\X{}$};
    
    \node [param, label = below:{ $\lambda$}] (lambda) at (4.5,-1.5) {};
    
    \node [param, label = below:{ $\nu$}] (nu) at (0.0,0.0) {};
    
    \node [RV,label={[xshift=-10pt,yshift=8pt]}] (f0) at (1.5,0.0) {\small$\fnull{}$};
    
    \node [RV,label={[xshift=-10pt,yshift=8pt]}] (theta) at (3.0,0.0) {$\theta$};
    
    \node [RV] (W) at (4.5,0.0) {\small$\W{}$};
    
    \node [RV] (fK) at (2.25,-1.5) {\small$\fK{}$};
    
    \node [RVo] (Y) at (2.25,-3.0) {\small$\Y{}$};
    
    \node [plate, inner xsep = 0.5cm, yshift=0.15cm, minimum height = 3cm, fit = (f0)(theta)(fK)(W)(nu)(lambda)] (plate1) {};
    \node[anchor=south east,inner sep=5pt] at (plate1.south east) 
    {$C$};
    
    \node[plate, fit =  (f0)(theta)(fK)(X)(Y) , inner xsep=0.5cm] (plate2) {};
    \node[anchor=south east,inner sep=5pt] at (plate2.south east) 
    {$N$};

    
    \draw[->, thick] (nu) edge (f0) ;
    \draw[->,thick] (X) edge (f0) ;
    \draw[->,thick] (X) edge (theta);
    \draw[->,thick] (lambda) edge (W) ;
    \draw[->,thick] (W) edge (theta);
    
    \draw[->,thick] (f0) edge (fK) ;
    \draw[->,thick] (theta) edge (fK) ;
    
    \draw[->,thick] (fK) edge (Y);

    \end{tikzpicture}

%% file: images/PGM_ETGP.tex
    \begin{tikzpicture}[baseline]
    
    \tikzstyle{RV}=[shape=circle,line width=1.0pt,minimum size = 0.75cm, draw=black,fill=gray!20]
    \tikzstyle{RVo}=[shape=circle,line width=1.0pt,minimum size = 0.75cm, draw=black,fill=white]
    \tikzstyle{param}=[fill,circle,inner sep=1.0pt,minimum size=0.5pt]
    \tikzstyle{plate}=[draw, rectangle, rounded corners]

    \node [RVo] (X) at (0.0,1.5) {\small$\X{}$};
    
    \node [RV,label={[xshift=-10pt,yshift=8pt]}] (f0) at (-1.5,1.5) {\small$\fnull{}$};
    
    \node [param, label = below:{$\nu$}] (nu) at (-3.0,1.5) {};
    
    \node [RV,label={[xshift=-10pt,yshift=8pt]}] (theta) at (0.0,0.0) {\small$\theta$};
    
    \node [RV] (W) at (1.5,0.0) {\small$\W{}$};
    
    \node [RV] (fK) at (-1.5,-1.5) {\small$\fK{}$};
    
    \node [param, label = below:{ $\lambda$}] (lambda) at (1.5,-1.5) {};

    \node [RVo] (Y) at (-1.5,-3.0) {\small$\Y{}$};

   \node [plate, inner xsep = 0.75cm, xshift=-0.1cm, yshift=0.15cm, minimum height = 3cm, fit = (theta)(fK)(W)(lambda)] (plate1) {};
   \node[anchor=south east,inner sep=5pt] at (plate1.south east) {$C$};
    
    \node[plate, fit =  (f0)(theta)(fK)(X)(Y) , inner xsep = 0.5cm] (plate2) {};
    \node[anchor=south east,inner sep=5pt] at (plate2.south east) {$N$};

    
    \draw[->, thick] (nu) edge (f0) ;
    \draw[->,thick] (X) edge (f0) ;
    \draw[->,thick] (X) edge (theta);
    \draw[->,thick] (lambda) edge (W) ;
    \draw[->,thick] (W) edge (theta);
    
    \draw[->,thick] (f0) edge (fK) ;
    \draw[->,thick] (theta) edge (fK) ;
    
    \draw[->,thick] (fK) edge (Y) ;

    \end{tikzpicture}

%% file: sections/C_Model_specification.tex
\section{Efficient Transformed Gaussian Processes}
In this section we describe the proposed method for multi-class
\GP classification, which we show is a more efficient application of the \TGP{} to classification problems with large $C$; with the additional benefit of naturally modeling dependencies between the processes. Our proposed method is specified by transforming a single sample from a \GP using $C$ invertible transformations, each of them mapping the \GP{} to a latent function for each class label. The generative procedure is given by:
\begin{align}
	\fnbnull{}(\cdot) &\sim \GP(0,K_\nu(\cdot,\cdot))\,, &
	\fnbK{1}(\cdot) &= \Gcompcinpdep{1}{\X{}}(\fnbnull{}(\cdot))\,, & \cdots &&  \fnbK{C}(\cdot) &= \Gcompcinpdep{C}{\X{}}(\fnbnull{}(\cdot)).
\end{align}
Using again the change of variable formula and inverse function theorem employed in \usec \ref{sec:background_tgp}, the joint conditional distribution of the $C$ processes is given by:
\begin{align}
\hspace{-0.25cm}p(\fallk|\Wall{})=p(\fnull{})\hspace{-0.1cm} \Jacobiancinputdepshortinline{\fpos{k}{1}}{1}{\Xsamples}\hspace{-0.1cm}\myprodinline{c= 2}{C} \mydeltafinpdepshort{c}{1}{\Xsamples}, \hspace{-0.1cm}
\label{jointETGPprior}
\end{align}
where $\delta(\cdot)$ denotes the Dirac measure and 
we define $\HH=\G^{-1}$ to be the corresponding inverse transformation.
Note that this decomposition is not unique. We label $\fK{1}$ as the \emph{pivot} and 
note that this joint distribution can be written equivalently \wrt any other 
\emph{pivot} $\fK{c}$ with $c \neq 1$.

This formulation has the advantage that a single \GP{} is used in practice which implies a constant scaling 
of \GP operations \wrt the number of classes,
speeding-up computations. This contrasts with a naive use of \TGP{}s or the 
model described in \usec \ref{sec:background_mgp} and motivates the name given to our model which we call \emph{Efficient} Transformed Gaussian Processes (\ETGP), particularly suited for multi-class problems 
with large $C$. Moreover, the resulting $C$ processes are dependent since they
share the same latent sample $\fnbnull{}$ from the original process. Also, since we use \NN{}s to parameterize the flows, the $C$ processes are 
non-stationary. \fig \ref{ETGPTGPPGM} compares the graphical models 
of the naive use of \TGP for multi-class classification 
and the proposed method \ETGP, highlighting their differences.

\begin{wrapfigure}{r}{0.4\textwidth}
    \centering
    \resizebox{0.4\textwidth}{!}{
    \input{./images/Nnet_arch2}}
\caption{Architecture of the NN used in this work. 
	Hidden layers are shared, with the final layer giving 
	each of the parameters of the $C$ flows.}
\label{fig:arch_nn}
\end{wrapfigure}
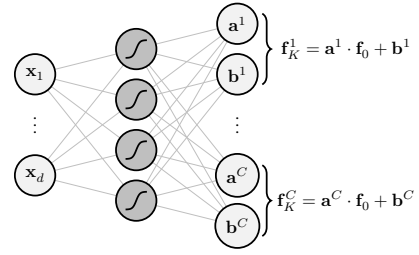

Since our work is focused on presenting \ETGP{}s as a good performance method at a low computational cost, we also propose an efficient \NN parameterization of the flow parameters, illustrated 
in \fig \ref{fig:arch_nn}. In principle, we could use a \NN per flow parameter, 
implemented efficiently using batched operations \citep{Blas_library}. However, 
by using a single \NN whose output layer coincides with the number of parameters 
per flow times the number of classes, we achieve a more efficient parameterization. 
For example, if we are modeling Imagenet ($C=1000$) with a linear flow 
($|\theta|=2$), then we would use an output layer of $2000$ neurons. Beyond 
being more efficient, this could also serve as a possible regularizer 
in the same fashion as sharing parameters regularizes \DNN{}s, something 
commonly used in computer vision applications through a back-bone convolutional model.

Another appealing property of the \ETGP is that we can efficiently create 
$C$ non-stationary dependent \GP{}s, by using a linear flow: 
$\fnbK{}{}(\X{})=a(\X{}) \fnbnull{}(\X{}) + b(\X{})$, as 
shown in \fig \ref{fig:arch_nn}. The following proposition characterizes the
corresponding joint prior distribution $p(\fallk{})$:
\begin{proposition}
    The joint conditional distribution of $C$ non-stationary \GP{}s obtained via a linear flow is given by:
    \begin{align}
         p(\fallk\mid\Wall{})=\Ngaussshort{\fK{1}\mid\mathbf{b}^1, \A{1} K_\nu(\Xsamples,\Xsamples) \A{1}^\text{T}} \myprodinline{c= 2}{C} \mydeltafinpdepshort{c}{1}{\Xsamples}	\,,
    \end{align}
    with $\mathbf{b}^1=(b^1(\X{1}),\ldots,b^1(\X{N}))^\text{T}$ 
	and $\A{1}\in \mathds{R}^{N\times N}$ a diagonal matrix with entries 
	$\aaa{1} = (a^1(\X{1}),\hdots,a^1(\X{N}))^\text{T}$. 
	Each marginal $p(\fK{c})$ is Gaussian with mean and covariance 
	given by $\mathbf{b}^c$ and $\A{c} K_\nu(\mathbf{X},\mathbf{X})\A{c}^\text{T}$, respectively. 
	The covariance matrix between the pivot $\fK{1}$ and $\fK{c}$, for $c\neq 1$, is given 
	by: $\aaa{1}{(\aaa{c})}^\text{T} \odot K_\nu(\Xsamples{},\Xsamples{})$ with $\odot$ denoting Hadamart product. Proof given in \uappendix \ref{math_appendix}.
    \label{prop1}
\end{proposition}

%% file: images/Nnet_arch2.tex
\tikzset{dist/.style={path picture= {
    \begin{scope}[x=1pt,y=10pt]
      \draw plot[domain=-6:6] (\x,{1/(1 + exp(-\x))-0.5});
    \end{scope}
    }
  }
}

\begin{tikzpicture}[node distance=2cm]
\tikzstyle{neuron_hidden}=[shape=circle,minimum width=0.8cm,draw=black,fill=gray!50, line width=1.0pt,dist]
\tikzstyle{neuron}=[shape=circle,minimum width=0.8cm,draw=black,fill=gray!10, line width=1.0pt]
\tikzstyle{dots}=[shape=circle]
\tikzstyle{write}=[]

\foreach \x in {0,...,3}
   \draw[gray!50,line width=0mm] (0.0,2.5) -- (2.0,\x);

\foreach \x in {0,...,3}
   \draw[gray!50,line width=0mm] (0.0,0.5) -- (2.0,\x);

\foreach \x in {0,...,3}
{
   \draw[gray!50,line width=0mm] (2.0,\x) -- (4.0,3.5);
   \draw[gray!50,line width=0mm] (2.0,\x) -- (4.0,2.5);
   \draw[gray!50,line width=0mm] (2.0,\x) -- (4.0,0.5);
   \draw[gray!50,line width=0mm] (2.0,\x) -- (4.0,-0.5);
}

\node [neuron] (n11)  at (0.0,2.5) {$\X{}_1$};
\node [write]  (n12)  at (0.0,1.6) {$\vdots$};
\node [neuron] (n13)  at (0.0,0.5) {$\X{}_d$};

\foreach \x in {0,...,3}
    \node [neuron_hidden] (n2\x) at (2.0,\x) {};

\node [neuron] (n41) at (4.0,3.5) {$\mathbf{a}^1$};
\node [neuron] (n42) at (4.0,2.5) {$\mathbf{b}^1$};
\node [write]  (n43) at (4.0,1.6) {$\vdots$};
\node [neuron] (n41) at (4.0,0.5) {$\mathbf{a}^C$};
\node [neuron] (n42) at (4.0,-0.5) {$\mathbf{b}^C$};

\draw [decorate,decoration={brace,amplitude=5pt,raise=4pt},xshift=10pt,line width=1.0pt]
(4.0,3.7) -- (4.0,2.3) node [black,midway,xshift=1.8cm] {$\fK{1}=\mathbf{a}^1\cdot\fnull{}+\mathbf{b}^1$};
\draw [decorate,decoration={brace,amplitude=5pt,raise=4pt},xshift=10pt,line width=1.0pt]
(4.0,0.7) -- (4.0,-0.7) node [black,midway,xshift=1.8cm] {$\fK{C}=\mathbf{a}^C\cdot\fnull{}+\mathbf{b}^C$};
\end{tikzpicture}

%% file: sections/D_inference.tex
\subsection{Approximate Inference}

Our inference algorithm is inspired by the key observations 
of \citep{VIinducingpoints_titsias,GPsBigData_hensman,TGP_maronas}. More precisely, we 
rely on a sparse \VI algorithm where the variational distribution is defined so that 
the conditional's model prior $p(\fK{} \mid \uK{})$ 
cancels \citep{VIinducingpoints_titsias}, without marginalizing out the process values at the inducing points to allow for 
mini-batch optimization \SVI \citep{GPsBigData_hensman}, and by defining the variational distribution over the \GP space and then warping it with the same flows as the prior \citep{TGP_maronas}.

To start with, a set of $M$ inducing points is defined on the \GP space $\fnbnull{}(\cdot)$. 
Note, however, that we can easily extend our framework to use inter domain inducing points 
\citep{interdomainGP_seminal} since we just need to derive the corresponding cross-covariances. 
Let $\uallk{}$ be defined as $\fallk{}$ in \ueqn \ref{jointETGPprior} summarizing the $C$ transformed process values at the inducing points $\Zsamples{}$.
In \uappendix \ref{math_appendix} we show that the joint conditional prior is:
\begin{align}
	& p(\fallk, \uallk|\Wall) = p(\fnull{}\mid\unull{}) \Jacobiancinputdepshortinline{\fpos{k}{1}}{1}{\Xsamples{}}\myprodinline{c= 2}{C} \mydeltafinpdepshort{c}{1}{\Xsamples}
	\nonumber \\
	& p(\unull{}) \Jacobiancinputdepshortinline{\upos{k}{1}}{1}{\Zsamples{}}  \myprodinline{c = 2}{C}\mydeltauinpdepshort{c}{1}{\Zsamples{}}\label{equ:sparse_joint}
\end{align}
This is possible because \ETGP is a consistent process, which means we can extend its finite 
dimensional distribution with inducing point locations as in standard \GP{}s. 
See \citep{TGP_maronas} for further details.

The variational distribution is assumed to have a similar form to the prior with some factors that are shared between them and others that are specific of the posterior approximation:
$q(\fallk,\uallk,\Wall)
=q(\fallk,\uallk|\Wall)q(\Wall)$ 
as in \citep{TGP_maronas}, where the variational distribution over the \NN{} has 
parameters $\phiall$ and is assumed to factorize across classes. Following 
\citep{VIinducingpoints_titsias,TGP_maronas}, the variational distribution over 
the values of the random processes at $\Xsamples{}$ and $\Zallsamples{}$, $q(\fallk,\uallk) = p(\fallk\mid\uallk)q(\uallk)$, is 
defined using the conditionals model's prior $p(\fallk\mid\uallk)$, given 
in \ueqn \ref{equ:sparse_joint}, and a free form variational 
distribution $q(\uallk)$. As in \citep{TGP_maronas}, $q(\uallk)$ is defined by warping a multivariate Gaussian defined on the original $\fnbnull{}$ space 
$q(\unull{}\mid\varm{},\varS{})$ using $\Gall$, 
where $\varm{} \in \mathds{R}^M,\,\varS{} \in \mathds{R}^{M\times M}$ are the
mean and covariance variational parameters:
\begin{align}
\hspace{-0.3cm} q(\uallk) = q(\unull{}) \Jacobiancinputdepshortinline{\upos{k}{1}}{1}{\Zsamples{}}\myprodinline{c= 2}{C} \mydeltauinpdepshort{c}{1}{\Zsamples{}}
\end{align}
The resulting \ELBO on the log-marginal-likelihood is, after several factor cancellations, equal to:
\begin{equation}
\begin{split}
\ELBO & = \mysuminline{n=1}{N}\mysuminline{c=1}{C} \mathds{I}(\Y{n}=c)\mathds{E}_{q(\fnbpos{0,n}{})q(\Wall)}[ \log \pi_c(\Gcompcinpdep{1}{\X{n}}(\fnbpos{0,n}{}),\hdots,\Gcompcinpdep{C}{\X{n}}(\fnbpos{0,n}{}))] \\ 
     & -\KLD[q(\unull{})||p(\unull{})] -  \mysuminline{c=1}{C}\KLD[q(\W{c})||p(\W{c}))]\hspace{-0.05cm}
\label{elbo}
\end{split}
\end{equation}
where $q(\fnbnull{})$ is computed as in \usec \ref{sec:background_mgp}. The expectation \wrt $q(\Wall)$ is computed via Monte Carlo and a single $1$-d quadrature is used to compute expectations over  $q(\fnbnull{})$. $\KLD[q(\unull{})||p(\unull{})]$ is tractable. $\KLD[q(\W{c})||p(\W{c}))]$ can be 
computed in closed-form for certain choices of the prior and variational posterior. However, 
we follow \citep{TGP_maronas} and use Monte Carlo Dropout \citep{MCdropout} (\MCDROP) rather 
than \VI to perform inference on $\W{}$. With this, the same NN
can be used to make non-Bayesian point estimate predictions 
(\PEETGP) \citep{dropouthinton} or Bayesian predictions (\BAETGP) \citep{MCdropout}.
This objective can be maximized using stochastic optimization methods and the
data can be sub-sampled for mini-batch training.
Readers concerned with the cancellation of delta functions in \ueqn \ref{elbo} can 
replace them with Gaussians with variance $\sigma^2$ to then take the limit $\sigma^2 \rightarrow 0$.

Predictions for $y^\star$ associated to a new $\mathbf{x}^\star$ are computed using 
an approximate predictive distribution: 
\begin{equation}
\textstyle p(y^\star\mid\X{\star},\Dsamples) \approx \mathds{E}_{q(\fnbnull{}(\X{\star}))q(\Wall)}[p(\Y{\star} \mid \Gcompcinpdep{1}{\X{\star}}(\fnbnull{}(\X{\star})),\hdots,\Gcompcinpdep{C}{\X{\star}}(\fnbnull{}(\X{\star})) )],
    \label{pos_predictive}
\end{equation}
where the integral is approximated by Monte Carlo and $1$-d quadrature having marginalized out $\uallk{}$.

\subsection{Summary of the proposed method and computational cost}

\ETGP creates $C$ dependent processes since $\fnull{}$ is shared. 
We have characterized the dependencies of these $C$ processes for a linear flow in 
\prop \ref{prop1}. The flows, however, need not be linear and can be arbitrarily complicated.
Because $\mathbb{G}_{\theta_K}^c$ is input-dependent the $C$ processes are also non-stationary.

Expectations \wrt the \NN{}'s parameters can be computed using batched matrix multiplications. Expectations \wrt $q(\fnull{})$ in \ueqn \ref{elbo} and \ueqn \ref{pos_predictive} can be computed with $1$-d quadrature. By contrast, the \SVGP{} method from \usec \ref{sec:background_mgp} cannot 
use quadrature methods. 
Moreover, the number of \GP{} operations is constant with $C$. To get  $q(\f{c})$ in \SVGP{}s \usec \ref{sec:background_mgp} one needs a cubic operation to invert $\GPcovar{c}(\Z{},\Z{})$ and $M^2$ operation to compute the variational parameters per class and datapoint, giving a complexity of $\complexity{CM^3 + CNM^2}$. This can be alleviated by 
sharing $K_\nu$ and $\Zsamples{}$ across \GP{}s, resulting in
$\complexity{M^3 + CNM^2}$, at the cost of limiting expressiveness, as we'll show. \ETGP cost is always $\complexity{M^3 + NM^2}$ (without considering 
the \NN'{}s computations, which for the architecture presented is often much faster and can be done in parallel to \GP{} operations).

%% file: sections/E_related_work.tex
\section{Related Work}

On the non-stationary side, the traditional approach is to use non-stationary covariance functions such as the Neural 
Network \citep{NNetKernelWill} or the Arcosine 
\citep{Arcoskernel}. Our experiments, however, show 
that \ETGP provides superior results in the multi-class setting when 
compared to a method using these kernels. One can also make stationary 
kernels non-stationary by making the parameters of the kernel depend 
on the input \citep{NScovheinonen16}. However, the work in \citep{NScovheinonen16} is 
limited to small datasets since it does not consider sparse \GP{}s and
it relies on Hamilton Monte Carlo for approximate inference, which is 
computationally expensive. Nevertheless, a sparse approach would require a \GP{} per kernel hyperparameter which can lead to a big number of \GP{}s for high $d$. Another approach to obtain non-stationary processes
considers stochastic processes mixing using hierarchical models \citep{KangruiNSNS} or by placing \GP{}s over the mixing matrix entries, achieving input-dependent length scales and 
amplitudes \citep{GPRNwilson}. These works either don't scale for high $C$ \citep{GPRNwilson} or require domain knowledge to avoid misspecification \citep{KangruiNSNS}. 

Non-stationarity can also be obtained by warping the input 
space using a non-linear transformation before introducing the data into the 
kernel \citep{sampson1992nonparametric,Schmidt00bayesianinference,manifold_gp_calandra:2016,deep_kernel_wilson:2016}. 
These methods, however, either run the risk of over-fitting the observed data, as 
a consequence of not regularizing the parameters of the non-linear transformation nor 
using a fully Bayesian approach, or either do not scale to large datasets. 

One can also use more sophisticated processes such as the 
\GP{} Product Model \citep{GPPM}, \DGP{}s \citep{DGP_seminal} or \TGP{}s 
\citep{TGP_maronas} to achieve non-stationarity. The \GP{} Product Model, however, does not 
scale to large datasets. \DGP{}s have been shown to give similar results to those 
of \TGP{}s. However, \TGP{}s have a lower computational cost than \DGP{}s and slightly higher than \SVGP{}s.  
Therefore, the proposed method, \ETGP, is expected to be faster than \TGP{}s and also faster than
\DGP{}s, in consequence.

On the dependence point of view, several approaches have been considered, which 
range from using process convolutions \citep{DependentGP} to mix latent 
\GP{}s via a mixing matrix \citep{LMCReview} whose entries can be parameterized 
by a \GP{} \citep{GPRNwilson}. More recently, \citep{JankowiakMOGP} extends 
Multi-output \GP{}s \citep{LMCReview}, Gaussian Process Regression Networks 
\citep{GPRNwilson} and \DGP \citep{DGP_seminal} by using \NN{}s to replace 
different building blocks of these methods. Since the computational 
cost of considering several \GP{}s for inference  is high 
(not necessarily for multi-class learning), 
several methods have tried to alleviate this cost by, \emph{e.g.}, using sparse methods 
\citep{SparseConvolvedGP} or more recently by assuming that the data 
lives around a linear subspace and then exploit a low-rank structure of 
the covariance matrix \citep{ScalableMOGPBruinsma}. All these works, however,
use several \GP{}s for modeling the data, unlike the proposed method \ETGP, 
and are hence expected to be more expensive.

%% file: sections/F_experiments.tex
\section{Experiments}

We evaluate \ETGP in $5$ \UCI datasets 
\citep{UCI_datasets} (see \fig \ref{fig:accuracy_stationary_mixing} for details). We compare \ETGP with \LINEAR, \SAL and \TANH flows with 
Bayesian (\BAETGP) and point estimate (\PEETGP) flow parameters predictions. 
We compare against a stationary independent (\RBF), as described in \usec \ref{sec:background_mgp}, and dependent (\RBFCORR) \SVGP{}s, where 
dependencies are obtained by mixing $C$ latent \GP{}s \citep{LMCReview}. 
We also compare against two non-stationary \SVGP 
with an arccosine (\ARCCOS) \citep{Arcoskernel} and a Neural Network (\NNET) \citep{NNetKernelWill} kernel. 
In \SVGP, we run the model with separate/shared kernels and inducing points 
across classes, indicated with separate/shared $K_\nu$ in the results. 
We report accuracy (\ACC) here. \uappendix \ref{experiments_appendix}
contains log-likelihood (\LL) results and gives all training details. We highlight that on each \SVGP{} run (one per training hyperparameters), we pick the best result on the \emph{test set}, so that the comparison with the \ETGP is the most pessimistic. By contrast, we perform model selection with a validation set for \ETGP. The code for \ETGP will be released in Github.

\subsection{Comparison against stationary dependent/independent \SVGP}
\label{subsec_stationary}

\begin{figure}[tbh!]
    \centering
    \includegraphics[width=\textwidth]{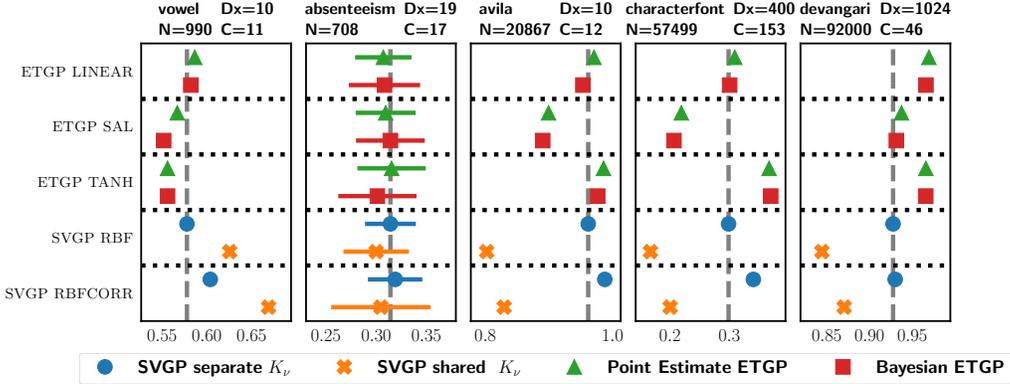}
    \caption{Avg. accuracy (right is better) comparing \ETGP vs. independent/dependent stationary \GP{}s. }
    \label{fig:accuracy_stationary_mixing}
\end{figure}

We compare \ETGP against stationary dependent/independent \SVGP{}s. The 
results obtained are displayed in \fig \ref{fig:accuracy_stationary_mixing}. We observe that on the big datasets with a large number of classes 
(\emph{i.e.}, \texttt{characterfont} and \texttt{devangari}) \ETGP clearly 
outperforms \SVGP{}s. In the worst case, the performance gain 
goes from $0.34 \,\RBFCORR$ to $0.36 \, \TANH$. In the best case,  
we see a boost from $0.29 \,\SVGP$ to $0.36 \,\TANH$ both in \texttt{characterfont}.
In \texttt{devangari} \ETGP{}s are clearly better in all cases boosting accuracy 
from $0.93 \,\RBFCORR$ to $0.96 \,\TANH \,\text{and} \,\LINEAR$, nearly matching the 
result obtained by a convolutional neural network ($0.98$) \citep{DevangariPaper}.
This boost in performance is obtained around one order of magnitude faster, see 
\usubsec \ref{timing_comparison} and \fig \ref{fig:train_time_comparison}.
We also see that sharing inducing points and covariances (orange crosses) clearly 
drops performance. In \texttt{avila}, a medium size dataset, \ETGP ($0.985\, \TANH$ $0.970\, \LINEAR$) 
works better than \SVGP ($0.962$) and comparable to correlated \SVGP ($0.988$), one order of magnitude faster. 

In the small datasets (\texttt{vowel} and \texttt{absenteeism}) we observe different 
things. First on \texttt{vowel} \SVGP sharing kernel and inducing points works best. 
This is because the training and test sets were collected from different speakers 
pronouncing vowels, and this domain shift can't be captured by these models. As 
a consequence, the shared $K_\nu$ model generalizes better since it under-fits
the training set (reflected by worse \ACC, \LL and \ELL in the training set), 
unlike separate $K_\nu$ and \ETGP. Since \ETGP chooses hyper-parameters using 
validation data extracted from the training set, this method cannot capture this domain 
shift.  However, in some runs, \ETGP (mostly those with higher dropout 
probability \ie higher epistemic uncertainty) was able to match the results of 
\SVGP with shared covariances. This remarks that epistemic uncertainty is 
beneficial in domain-shift small datasets, as expected. We emphasize that this is not a problem 
of the \ETGP{} but derived from the characteristics of the dataset itself since 
it is also suffered by \SVGP with separate $K_\nu$. Note that on 
\texttt{absenteeism} (fewer training points than \texttt{vowel}) we 
see that the proposed model ($0.307\, \LINEAR$ and $0.315\,\TANH$ ) performs 
similarly to \SVGP ($0.314$) and correlated \SVGP ($0.319$).
Finally, across all datasets, the $\SAL$ flow is the worst 
one (see \uappendix \ref{experiments_appendix} for possible explanations). We remark 
how well the $C$ non-stationary dependent \GP{}s (\LINEAR) perform here, 
opening its use in other \GP{} applications. 

Regarding \LL (see  \uappendix 
\ref{experiments_appendix}), we observe similar results across all datasets.  For small datasets \BAETGP provides much better uncertainty quantification (\LL), something we don't observe on the medium/large datasets in neither \ACC or \LL. This matches findings from \citep{TGP_maronas} where being Bayesian does not show improvements on classification datasets. This is expected as with big $N$ epistemic uncertainty vanishes, suggesting that alternatives that fix the dropout probability depending on the number of training points, such as concrete dropout \citep{ConcreteDropout}, are a potential line of research to enhance the proposed model.
\begin{figure}[tbh!]
    \centering
    \includegraphics[width=\textwidth]{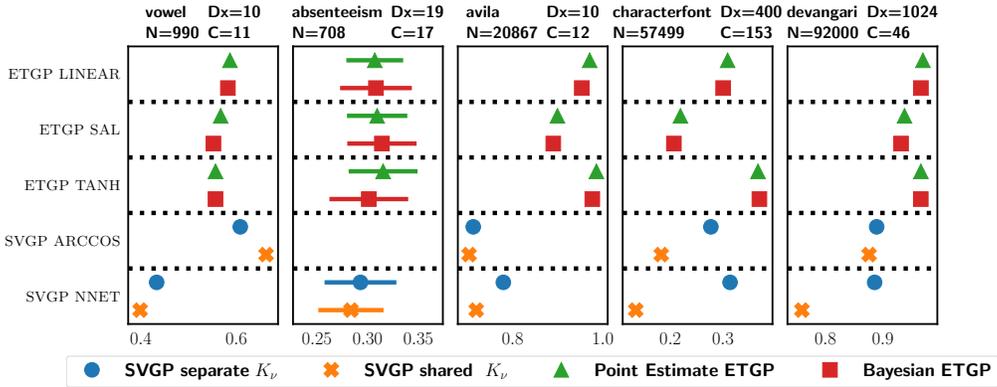}
    \caption{Avg. accuracy (right is better) comparing \ETGP with two non-stationary \GP{}s. }
    \label{fig:accuracy_non-stationary}
	\vspace{-.25cm}
\end{figure}
\subsection{Comparison against Non-Stationary \GP{} priors}
\label{subsec_non_stationary}
We compare \ETGP also against two non-stationary kernels \fig
 \ref{fig:accuracy_non-stationary}.  We observe that the \NNET kernel often gives worse results 
than the \ARCCOS or \ETGP, especially for shared kernels across classes. This remarks the importance 
of having background knowledge about the non-stationarity of the particular application. Our work and \citep{TGP_maronas} 
show that the non-stationarity achieved by input dependent flows is beneficial and easy to
interpret since we just make each of the marginals depend directly on the part of 
the feature space that we are modeling, with no cross interactions between data points 
beyond those given by the base stationary kernel. \ETGP outperforms both the 
\NNET and \ARCCOS kernel. In \texttt{vowel}, the shared \SVGP kernel works the 
best, matching the results of \usec \ref{subsec_stationary}. In terms of \LL 
our model also works consistently better. See \uappendix \ref{experiments_appendix} for details. We don't show results for \ARCCOS on \texttt{absenteeism} as we found training runs to saturate numerically (using float64 precision).
\subsection{Timing comparison}
\label{timing_comparison}

\begin{figure}[tbh!]
    \centering
    \includegraphics[width=\textwidth]{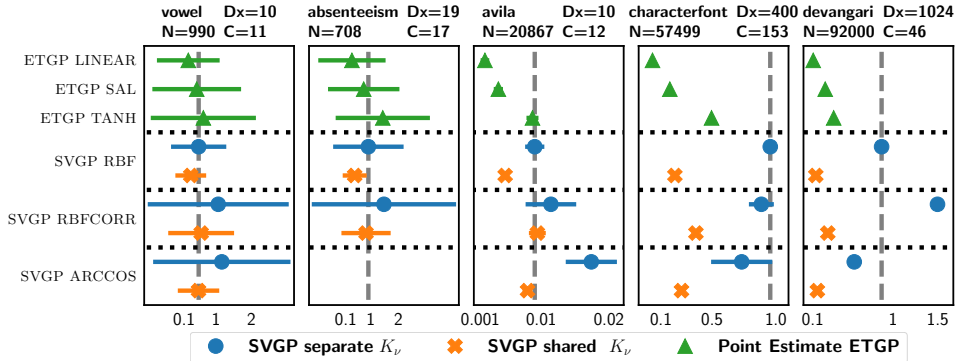}
    \caption{Average training time per epoch in minutes
	(left is better) comparing \ETGP{} with \SVGP{}s. 
	\NNET kernel is omitted as it is slower. 
	Times for \texttt{absenteeism} and \texttt{vowel} are scaled by $10^3$. }
    \label{fig:train_time_comparison}
	\vspace{-.25cm}
\end{figure}

We report the average training time in \fig \ref{fig:train_time_comparison} and 
prediction time in \uappendix \ref{experiments_appendix}. 
By using \MCDROP, the training time of the \PEETGP and \BAETGP is the same. Predictions 
for the Bayesian \ETGP{}s can be computed in parallel. We 
refactorize \GPFLOW{}'s source code so that the shared \SVGP model is more efficient, see \uappendix \ref{gpflow_source_code_refactorization}. We
observe that \ETGP is the fastest method, with a gain of 
one order of magnitude, compared to \SVGP. \SVGP with shared 
$K_\nu$ is competitive in terms of training time, 
but has a drop in performance (see \usubsec \ref{subsec_stationary}),
unlike \ETGP which typically performs best.

\subsection{Fewer inducing points act as a regularizer}

\begin{figure}[tbh!]
    \centering
    \includegraphics[width=\textwidth]{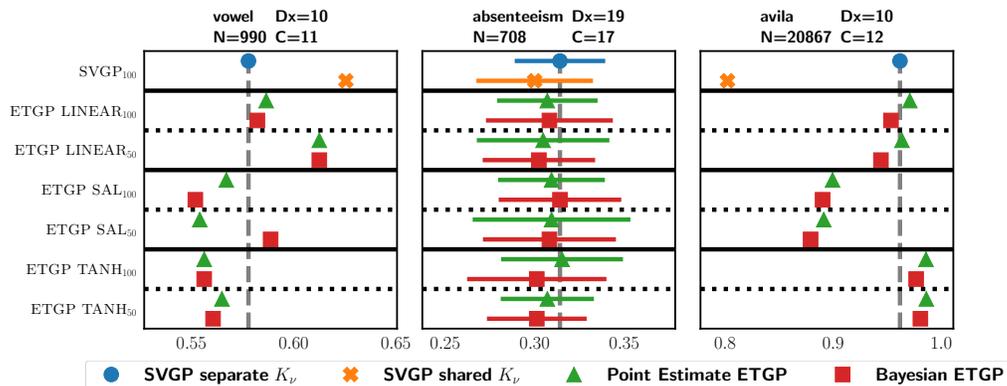}
    \caption{Results comparing \ACC (right is better) using less inducing points.}
    \label{fig:less_inducings}
	\vspace{-.25cm}
\end{figure}

In \citep{TGP_maronas} it is showed that \TGP{}s could match 
\SVGP{}s performance in regression problems using $20$ times less inducing 
points. We additionally found that using fewer inducing points can serve 
also as a regularizer, since the \GP{} posterior is expected to parameterize smoother 
functions in that case. With this goal, we report results for \ETGP using $50$ and $100$ inducing 
points, see \fig \ref{fig:less_inducings}. We observed that 
\ETGP gets regularized when using fewer inducing 
points. In \texttt{vowel} it improves results, and matches \SVGP{}s performance
in \texttt{avila} and \texttt{absenteeism}.
We extend this analysis in \uappendix \ref{experiments_appendix}.

%% file: sections/G_conclusions_future_work.tex
\section{Conclusions and future work}

\looseness=-2 We have introduced the Efficient Transformed Gaussian Process (\ETGP), as a way of 
creating $C$ dependent non-stationary stochastic processes in an efficient way. 
For this, a single initial \GP is transformed $C$ times. This has the benefit
of reducing the computational cost while providing enough model flexibility to
learn complex tasks. We have provided an efficient training algorithm for \ETGP based 
on variational inference. This method has been evaluated in the context of multi-class 
classification. Our results show that \ETGP{} is competitive or even better than 
typical sparse \SVGP{}s, at a lower computational cost. A limitation of \ETGP{} is, 
however, that it leads to a model that is more difficult to interpret than \SVGP{}s
as a consequence of the non-linear transformations, although some flows allow controlling the moments of the induced distributions \citep{compositionally_warped_gp_rios:2019}.
Future work can focus 
on extending \ETGP to multi-task and multi-label problems, where a large 
number of processes is also needed, and its use in a deep kernel learning 
framework \citep{deep_kernel_wilson:2016}. Also, an analysis of how the dependencies are modeled by sharing the copula of the base \GP{}, including how additional dependencies can be gained by mixing \ETGP{}s with a mixing matrix. The use of concrete dropout and using \GP{}s instead of \NN{}s to parameterize the flows are alternatives to improve different aspects of the Bayesian nature of the flow such as computational performance, well-specified Bayesian priors and epistemic uncertainty. 

%% file: sections/Z011_societal_impact.tex
\section{Societal Impacts}

This work introduces a new probabilistic machine learning method that can be used to make predictions quantifying its uncertainty. These uncertainty quantification make all these probabilistic methods particularly suited to be deployed in high risk scenarios such as autonomous driving or medicine. In all these problems we require the model to be interpretable and to provide reliable predictions. As a consequence, a careful analysis of the correct modeling of these properties must be done before deploying the model to avoid potential negative societal impacts.

Also, one needs to consider the particular limitations of the approximate inference algorithm and model assumptions on each particular application. For example the presented inference algorithm can only be used with diagonal flows, which leaves the dependencies (copula) of the \GP{} unchanged. Thus, if one believes the copula is non-Gaussian, deploying the \ETGP can result in biased predictions that can have harmful consequences.

%% file: sections/Z01_acks.tex
\acksection

The authors acknowledges funding coming from the Spanish National Research Project PID2019-106827GB-I00.  We also acknowledge the use of the computational resources from Centro de Computacion Cientifica (\acro{\smaller CCC}) and the \acro{\smaller AUDIAS} Laboratory both at Universidad Autónoma de Madrid. Part of this work was carried out while J.M. was at the \acro{\smaller PRHLT} research center at Universidad Politécnica de Valencia.

%% file: sections/Z0_ethics.tex
\clearpage
\section*{Checklist}

\begin{enumerate}

\item For all authors...
\begin{enumerate}
  \item Do the main claims made in the abstract and introduction accurately reflect the paper's contributions and scope?
    \answerYes{}
  \item Did you describe the limitations of your work?
    \answerYes{In the experimental section and the conclusions section we raise some of the issues of the proposed method. In general, our method provides advantages over baseline and we don't really find any problem beyond those inherited from the TGP.}
  \item Did you discuss any potential negative societal impacts of your work?
    \answerYes{This information can be found in the supplementary material}
  \item Have you read the ethics review guidelines and ensured that your paper conforms to them?
    \answerYes{Our algorithm is a general purpose classification algorithm which has additional features like non-stationarity or being more efficient. Thus it does not raise any additional ethical problem compared to standard classification.}
\end{enumerate}

\item If you are including theoretical results...
\begin{enumerate}
  \item Did you state the full set of assumptions of all theoretical results?
    \answerYes{See proposition 1}
        \item Did you include complete proofs of all theoretical results?
    \answerYes{See appendix where we demonstrate}
\end{enumerate}

\item If you ran experiments...
\begin{enumerate}
  \item Did you include the code, data, and instructions needed to reproduce the main experimental results (either in the supplemental material or as a URL)?
    \answerYes{Yes, the code is in the supplementary material}
  \item Did you specify all the training details (e.g., data splits, hyperparameters, how they were chosen)?
    \answerYes{We do it in the supplementary material}
        \item Did you report error bars (e.g., with respect to the random seed after running experiments multiple times)?
    \answerYes{In all cases.}
        \item Did you include the total amount of compute and the type of resources used (e.g., type of GPUs, internal cluster, or cloud provider)?
    \answerYes{We use an homogeneous cluster to launch our experiments.}
\end{enumerate}

\item If you are using existing assets (e.g., code, data, models) or curating/releasing new assets...
\begin{enumerate}
  \item If your work uses existing assets, did you cite the creators?
    \answerYes{We cite GPflow or UCI}
  \item Did you mention the license of the assets?
    \answerNo{That is already given in the Github of the corresponding codebase}
  \item Did you include any new assets either in the supplemental material or as a URL?
    \answerYes{We release our code in the supplementary material}
  \item Did you discuss whether and how consent was obtained from people whose data you're using/curating?
    \answerYes{All people submitting to UCI repository give consent to use the data. We properly cite UCI.}
  \item Did you discuss whether the data you are using/curating contains personally identifiable information or offensive content?
    \answerNA{The datasets used do not have personally identifiable information}
\end{enumerate}

\item If you used crowdsourcing or conducted research with human subjects...
\begin{enumerate}
  \item Did you include the full text of instructions given to participants and screenshots, if applicable?
    \answerNA{We don't conduct research with human subjects...}
  \item Did you describe any potential participant risks, with links to Institutional Review Board (IRB) approvals, if applicable?
    \answerNA{We don't conduct research with human subjects...}
  \item Did you include the estimated hourly wage paid to participants and the total amount spent on participant compensation?
    \answerNA{We don't conduct research with human subjects...}
\end{enumerate}

\end{enumerate}

%% file: sections/Z1_equations_appendix.tex
\section{Math Appendix}
\label{math_appendix}

\ifapendixsubmission
{This information is attached in the supplementary  material.}
\else
{

In this appendix we provide a wider description of the equations involved in this paper, ranging from the model definition to the sparse variational inference algorithm. For completeness, and to make the improvements of the proposed model clearer, we start by describing the sparse variational inference algorithm applied to multi-class problems with and independent \GP{} prior, previous to the definition of our model. Although we compare against correlated \GP{}s using a mixing matrix, we don't provide its derivation since it goes beyond the scope of this appendix. In the end of this appendix we proof \prop \ref{prop1}. If possible we keep all the conditioning set explicit in the equations presented.

\subsection{ Classification with independent \GP{} priors }

Given a classification problem with inputs $\X{}\in \Xspace \subseteq \mathds{R}^d$ and $C$ outputs $\Y{}\in \Yspace \subset \mathds{N}$, we want to learn $C$ mapping functions from $\X{}$ to the probability of belonging to each $\Y{}$; given a set of observations $\Dsamples=\left\{\X{n},\Y{n}\right\}^N_{n=1}$ with $\Xsamples=(\X{1},\ldots,\X{N})$ and $\Ysamples=(\Y{1},\ldots,\Y{N})$. 

If this modeling procedure is done using \GP{}s, then we place an independent \GP{} on each of these $C$ functions, each one parameterized by a mean function $\mu_\nu(\X{})$ (which we assume to be zero) and a covariance matrix $\GPcovar{c}(\X{},\X{})$, parameterized by $\nu$. Following the notation introduced in the main paper, the joint distribution of $C$ processes at locations $\Xsamples{}$ is given by:
\begin{align}
        p(\fallnull\mid\nuall) = \myprod{c=1}{C} \Ngauss{ \fnull{c} \mid \veczero, \GPcovar{c}(\Xsamples{},\Xsamples{})}
\end{align}
where $\fallnull=\{\fnull{1},\hdots,\fnull{C}\}$. This prior is combined with a likelihood $p(\Ysamples{}\mid\fallnull)$ that links latent functions to observations. In classification, a common choice is the Categorical Likelihood, giving the joint distribution:
\begin{align}
    p(\Ysamples{},\fallnull{}) & = 
	p(\Ysamples{}\mid\fallnull{}) p(\fallnull{}) = 
	\left[ 
	\myprod{n=1}{N} \myprod{c=1}{C} \pi_c(\fnballpos{0,n}{})^{\mathbb{I}(\Y{n}=c)} \right]
	\myprod{c=1}{C}\Ngauss{\fnull{c} \mid \mathbf{0} , \GPcovar{c}(\Xsamples{},\Xsamples{})}\,,
\end{align}
where $\pi_c(\fnballpos{0,n}{}) = \exp(\fnbnull{c}(\X{n}) / \mysuminline{c'=1}{C} \exp(\fnbnull{c'}(\X{n}))$ is the Softmax link function mapping latent vectors to probabilities, and $\mathbb{I}(\cdot)$ the indicator function.

In Bayesian learning, we are interested in the posterior $p(\fnull{}\mid\Dsamples)$, which is intractable for many likelihoods, and computationally unfeasible since its complexity scales cubically with the number of training points. We now introduce the variational sparse derivation.
\subsubsection{Variational sparse derivation}
The idea behind sparse \GP{}s is to use a set of $M$ inducing points $\Z{}\in\Xspace$, with $\Zsamples{}=(\Z{1},\ldots,\Z{M})$, that acts as summary statistics of the data. Each inducing point $\Z{}$ has an associated latent value $\unbnull{}$. Following the \GP{}'s prior definition, then at $\Zsamples{}$  we have $\unull{} \sim \GP(\unull{}\mid0,\GPcovar{}(\Zsamples{},\Zsamples{})))$. Thus, given $\Xsamples{},\Zsamples{}$ the joint distribution is Gaussian given by:

\begin{align}
p(\fnull{},\unull{}\mid\Xsamples{},\Zsamples{},\nu) = \Ngauss{
\begin{array}{c|c}
\fnull{} & \veczero  \\
\unull{} & \veczero  \\
\end{array},\hspace{-0.5cm} 
\begin{array}{cc}
     &  \GPcovar{}(\Xsamples{},\Xsamples{}),\,\, \GPcovar{}(\Xsamples{},\Zsamples{}) \\
     &  \GPcovar{}(\Zsamples{},\Xsamples{}),\,\, \GPcovar{}(\Zsamples{},\Zsamples{}) 
\end{array}
}
\end{align}
One key contribution from \citep{VIinducingpoints_titsias} is to define these inducing points to be variational parameters that are learned by minimizing the \KLD between the approximate $q(\fnull{},\unull{})$ and true posterior $p(\fnull{},\unull{}\mid\Dsamples,\Zsamples{},\nu)$. Since $\Z{}$ do not belong to the model parameters, then they don't increase the model expressiveness hence protecting the learning procedure from overfitting. 

The other key contribution from \citep{VIinducingpoints_titsias} is how the variational distribution is defined. In particular, $q(\fnull{},\unull{})=p(\fnull{}\mid\unull{})q(\unull{})$ so that the conditional's model prior $p(\fnull{}\mid\unull{})$ gets canceled. Later, \citep{GPsBigData_hensman} propose to keep $q(\unull{}\mid\varm{},\varS{})$ explicit by parameterizing it with a Gaussian distribution with parameters $\varm{} \in \mathbb{R}^M,\,\varS{} \in \mathbb{R}^{M\times M}$. With this, the \ELBO can be optimized with stochastic variational inference.

If $C$ independent \GP{}s are going to be used, we can easily extend the presented equations as follows. The joint prior factorizes across $C$ as above:
\begin{align}
p(\fallnull{},\uallnull{}\mid\Xsamples{},\Zallsamples{},\nuall) = \myprod{c=1}{C}\Ngauss{
\begin{array}{c|c}
\fnull{c} & \veczero  \\
\unull{c} & \veczero  \\
\end{array},\hspace{-0.5cm} 
\begin{array}{cc}
     &  \GPcovar{c}(\Xsamples{},\Xsamples{}),\,\, \GPcovar{}(\Xsamples{},\Zsamples{c}) \\
     &  \GPcovar{c}(\Zsamples{c},\Xsamples{}),\,\, \GPcovar{}(\Zsamples{c},\Zsamples{c})
\end{array}
}
\end{align}
and the approximate posterior can be defined by factorizing across $C$ as well:
\begin{equation}
    q(\fallnull{},\uallnull{}\mid\Xsamples{},\Zallsamples{},\nuall,\varmall,\varSall)=\myprod{c=1}{C}p(\fnull{c}\mid\unull{c},\Xsamples{},\Zsamples{c},\nu_c)q(\unull{c}\mid\varm{c},\varS{c})
\end{equation}
The \KLD minimization between the approximate posterior and the true posterior is equivalent to maximizing the Evidence Lower Bound (\ELBO), which we now derive:
\begin{equation}
\begin{split}
    \ELBO &= \int_{\fnull{1}} \hdots \int_{\fnull{C}}\int_{\unull{1}} \hdots \int_{\unull{C}} q(\fallnull{},\uallnull{}) \log \frac{p(\Ysamples{}\mid\fallnull{})p(\fallnull{},\uallnull{})}{q(\fallnull{},\uallnull{})}\dd\fnull{1}\hdots\dd\fnull{C}\dd\unull{1}\hdots\dd\unull{C}\\
    &=
    \underbrace{\int_{\fnull{1}} \hdots \int_{\fnull{C}}\int_{\unull{1}} \hdots \int_{\unull{C}} q(\fallnull{},\uallnull{}) \log p(\Ysamples{}\mid\fallnull{})\dd\fnull{1}\hdots\dd\fnull{C}\dd\unull{1}\hdots\dd\unull{C}}_{\ELL} \\
    & \underbrace{+ \int_{\fnull{1}} \hdots \int_{\fnull{C}}\int_{\unull{1}} \hdots \int_{\unull{C}} q(\fallnull{},\uallnull{}) \log \frac{p(\fallnull{},\uallnull{})}{q(\fallnull{},\uallnull{})}\dd\fnull{1}\hdots\dd\fnull{C}\dd\unull{1}\hdots\dd\unull{C}}_{-\KLD}
\end{split}
\end{equation}
where we have dropped the conditioning set for clarity. We now workout each term separately:
\begin{equation}
\begin{split}
   -\KLD &=  \int_{\fnull{1}} \hdots \int_{\fnull{C}}\int_{\unull{1}} \hdots \int_{\unull{C}} q(\fallnull{},\uallnull{}) \log\frac{p(\fallnull{},\uallnull{})}{q(\fallnull{},\uallnull{})}\dd\fnull{1}\hdots\dd\fnull{C}\dd\unull{1}\hdots\dd\unull{C} \\
   &= \int_{\unull{1}} \hdots \int_{\unull{C}} q(\uallnull{}) \log\frac{\cancel{p(\fallnull{}\mid\uallnull{})}p(\uallnull{})}{\cancel{p(\fallnull{}\mid \uallnull{})}q(\uallnull{})}\dd\unull{1}\hdots\dd\unull{C} \\
   &= \int_{\unull{1}} \hdots \int_{\unull{C}} \myprod{c=1}{C} q(\unull{c}) \mysum{c'=1}{C}\log\frac{p(\unull{c'})}{q(\unull{c'})}\dd\unull{1}\hdots\dd\unull{C} = \\
   &= \mysum{c'=1}{C}\int_{\unull{1}} \hdots \int_{\unull{C}} \myprod{c=1}{C} q(\unull{c}) \log\frac{p(\unull{c'})}{q(\unull{c'})}\dd\unull{1}\hdots\dd\unull{C} = \\
   &= \mysum{c=1}{C}\int_{\unull{c}}  q(\unull{c}) \log\frac{p(\unull{c})}{q(\unull{c})}\dd\unull{c} \\
   &= - \mysum{c=1}{C}\KLD[q(\unull{c})|| p(\unull{c})]
\end{split}
\end{equation}
The Expected log likelihood (\ELL) is given by:
\begin{equation}
    \begin{split}
        \ELL  &= \int_{\fnull{1}} \hdots \int_{\fnull{C}}\int_{\unull{1}} \hdots \int_{\unull{C}} q(\fallnull{},\uallnull{}) \log p(\Ysamples{}\mid\fallnull{})\dd\fnull{1}\hdots\dd\fnull{C}\dd\unull{1}\hdots\dd\unull{C} \\
        &=\int_{\fnull{1}} \hdots \int_{\fnull{C}} q(\fallnull{}) \log \myprod{n=1}{N}p(\Y{n}\mid\fnballpos{0,n}{})\dd\fnull{1}\hdots\dd\fnull{C} \\
        &= \int_{\fnull{1}} \hdots \int_{\fnull{C}} q(\fnull{1}) \hdots q(\fnull{C}) \log \myprod{n=1}{N}\myprod{c=1}{C} \pi_c(\fnballpos{0,n}{})^{\mathbb{I}(\Y{n}=c)}\dd\fnull{1}\hdots\dd\fnull{C} \\
        &= \mysum{n=1}{N}  \mysum{c=1}{C} \mathbb{I}(\Y{n}=c)\int_{\fnbpos{0,n}{1}} \hdots \int_{\fnbpos{0,n}{C}} q(\fnbpos{0,n}{1}) \hdots q(\fnbpos{0,n}{C}) \log \pi_c(\fnballpos{0,n}{})\dd\fnbpos{0,n}{1}\hdots\dd\fnbpos{0,n}{C} \\
    \end{split}
\end{equation}
recovering the bound of the main paper (\ueqn \ref{elbo_multiclass_independent_gp}):
\begin{equation}
    \begin{split}
        \ELBO &=      \mysum{n=1}{N}  \mysum{c=1}{C} \mathbb{I}(\Y{n}=c)\int_{\fnbpos{0,n}{1}} \hdots \int_{\fnbpos{0,n}{C}} q(\fnbpos{0,n}{1}) \hdots q(\fnbpos{0,n}{C}) \log \pi_c(\fnballpos{0,n}{})\dd\fnbpos{0,n}{1}\hdots\dd\fnbpos{0,n}{C} \\
              & -  \mysum{c=1}{C}\KLD[q(\unull{c})|| p(\unull{c})]
    \end{split}
\end{equation}

Note that this bound is amenable to stochastic optimization using minibatches, where the integrals are approximated by Monte Carlo using reparameterized gradients (\aka path-wise gradients). The \KLD can be computed in closed form.  Most importantly, each  $q(\fnbpos{0,n}{c})$ is a univariate Gaussian distribution given by:
\begin{equation}
\begin{split}
     q(\fnbpos{0,n}{c}) = \mathcal{N}( \fnbpos{0,n}{c} \mid & \GPcovar{c}{_{\X{n},\Zsamples{c}}} \GPcovar{c}{_{\Zsamples{c},\Zsamples{c}}}^{-1}\varm{c} ,  \\
     &  \GPcovar{c}{_{\X{n},\X{n}}}-\GPcovar{c}{_{\X{n},\Zsamples{c}}}\GPcovar{c}{_{\Zsamples{c},\Zsamples{c}}}^{-1}[\GPcovar{c}{_{\Zsamples{c},\Zsamples{c}}}+\varS{c}]\GPcovar{c}{_{\Zsamples{c},\Zsamples{c}}}^{-1}\GPcovar{c}{_{\Zsamples{c},\X{n}}} ) 
\end{split}\label{svgp_qf0}
\end{equation}
obtained by solving $\int_{\unull{1}}\hdots\int_{\unull{C}} \myprod{c=1}{C}p(\fnull{c}\mid\unull{c},\Xsamples{},\Zsamples{c},\nu_c)q(\unull{c}\mid\varm{c},\varS{c}) \dd\unull{1}\hdots\dd\unull{C} = \myprod{c=1}{C} \int_{\unull{c}} p(\fnull{c}\mid\unull{c},\Xsamples{},\Zsamples{c},\nu_c)q(\unull{c}\mid\varm{c},\varS{c})\dd\unull{c}$.
This computation needs to be computed $C$ times requiring a complexity of $\complexity{CM^3 + CNM^2}$, which can be reduced to $\complexity{M^3 + CNM^2}$ if the inducing points are shared. We can gain additional performance if the kernel is shared as noted in \uappendix \ref{gpflow_source_code_refactorization}. However, as shown in the experiments sharing kernel and inducing points can drop performance. 
\subsection{Classification with Efficient Transformed Gaussian Processes}

We now present the derivations required for the proposed model. This model is specified by transforming a single sample from a \GP{} using $C$ invertible transformations $\Gallcomp$ by the following generative procedure:
\begin{equation}
    \begin{split}
        	\fnbnull{}(\cdot) &\sim \GP(0,K_\nu(\cdot,\cdot))\\
	\fnbK{1}(\cdot) = \Gcompcinpdep{1}{\Xsamples}(\fnbnull{}(\cdot)),\,\,\,   &\cdots \,\,\,    \fnbK{C}(\cdot) = \Gcompcinpdep{C}{\Xsamples}(\fnbnull{}(\cdot)).
    \end{split}
\end{equation}
The prior distribution over $C$ processes is derived as follows. For exemplification purposes consider $C=3$ and consider the processes evaluation at the index set $\Xsamples$, then we have:
\begin{equation}
\begin{aligned}
    \hspace{-1.0cm} &\fnull{} \hspace{-0.4cm} &&\sim p(\fnull{}\mid\Xsamples{},\nu) \\
    \hspace{-1.0cm}&\fK{1}   \hspace{-0.4cm}&&= \Gcompcinpdep{1}{\Xsamples}(\fnull{});\,\, \fK{1} = \Gcdirinvcompinpdep{1}{2}{\Xsamples}(\fK{2});\,\, \fK{1} = \Gcdirinvcompinpdep{1}{3}{\Xsamples}(\fK{3})\\
   \hspace{-1.0cm} &\fK{2}   \hspace{-0.4cm}&&= \Gcompcinpdep{2}{\Xsamples}(\fnull{});\,\, \fK{2} = \Gcdirinvcompinpdep{2}{1}{\Xsamples}(\fK{1});\,\, \fK{2} = \Gcdirinvcompinpdep{2}{3}{\Xsamples}(\fK{3})\\
   \hspace{-1.0cm} &\fK{3}   \hspace{-0.4cm}&&= \Gcompcinpdep{3}{\Xsamples}(\fnull{});\,\, \fK{3} = \Gcdirinvcompinpdep{3}{1}{\Xsamples}(\fK{1});\,\, \fK{3} = \Gcdirinvcompinpdep{3}{2}{\Xsamples}(\fK{2})\\
\end{aligned}\label{example_derivation_CI_ETGP}
\end{equation}
with $\mathbb{H}\coloneqq\mathbb{G}^{-1}$. In order to define the prior probability of the classes we first observe that the following conditional independence holds from the construction introduced above:
\begin{equation}
    \begin{split}
        & p(\fK{1},\fK{2},\fK{3}) = p(\fK{1}) p(\fK{2}\mid\fK{1})p(\fK{3}\mid\fK{1},\cancel{\fK{2}})\\
        & p(\fK{2},\fK{1},\fK{3}) = p(\fK{2}) p(\fK{1}\mid\fK{2})p(\fK{3}\mid\fK{2},\cancel{\fK{1}})\\
        & p(\fK{3},\fK{1},\fK{2}) = p(\fK{3}) p(\fK{1}\mid\fK{3})p(\fK{2}\mid\fK{3},\cancel{\fK{1}})
    \end{split}
\end{equation}
where we have chosen to write the $3$ out of $6$ possibilities just for exemplification purposes.
This conditional independence holds because the probability of  $\fK{3}$ given $\fK{1},\fK{2}$  is given by a direct mapping either from $\fK{1}$ or $\fK{2}$ as illustrated in \ueqn \ref{example_derivation_CI_ETGP}. We define the \emph{pivot} to be the member on which we \emph{always} condition, \ie  if $p(\fK{3}\mid\fK{2},\fK{1}) = p(\fK{3}\mid\fK{1}) $ then the \emph{pivot} is $\fK{1}$. Note, however, that it will also be valid to choose any other member as a \emph{pivot}, for example $p(\fK{3}\mid\fK{2},\fK{1}) = p(\fK{3}\mid\fK{2})$. Finally, note that we can write the conditional distribution $p(\fK{3}\mid\fK{1})$ as:
\begin{equation}
p(\fK{3}\mid\fK{1}) = \mydeltafinpdep{3}{1}{\Xsamples}    
\end{equation}
with $\delta$ being the Dirac delta measure. Using both observations we can write the prior joint conditional distribution over the classes as:
\begin{equation}
\begin{split}
    p(\fallk\mid \Gallcomp,\Wall,\Xsamples{},\nu) =&\, p(\fnull{}\mid \Xsamples{},\nu)\Jacobiancinputdep{\fK{1}}{1}{\Xsamples} \\
    & \myprod{c=  2}{C} \mydeltafinpdep{c}{1}{\Xsamples}
\end{split}
\end{equation}
recovering the expression in \ueqn \ref{jointETGPprior} in the main paper. The overall joint is given by:
\begin{equation}
    p(\fallk,\Wall\mid\Gallcomp,\lambdaall,\Xsamples,\nu) = p(\fallk\mid\Gallcomp,\Wall,\Xsamples{},\nu) p(\Wall\mid\lambdaall)
\end{equation}
with $p(\Wall\mid\lambdaall)$ denoting the prior over the parameters of the Bayesian Neural Network (\BNN).

\subsubsection{Prior conditional distribution $p(\fallk\mid\uallk)$}

We now derive the prior conditional distribution $p(\fallk\mid \uallk)$. In a similar vein to \GP{}s we will derive a sparse variational inference algorithm, from where inducing points need to be incorporated. Note that since we use diagonal flows, the resulting joint distribution is consistent (\ie is a finite dimensional realization of a stochastic process), which means we can extend its index set introducing inducing points $\uallk$ at inducing locations $\Zsamples{}$, similar to what we do in \GP{}s.

First, note that following the previous section we can write the marginal distribution at the inducing points by:
\begin{equation}
\begin{split}
    p(\uallk \mid \Gallcomp,\Wall,\Zsamples{},\nu) =&\, p(\unull{}\mid \Zsamples{},\nu)\Jacobiancinputdep{\uK{1}}{1}{\Zsamples{}} \\
    & \myprod{c=  2}{C} \mydeltauinpdep{c}{1}{\Zsamples{}}.
\end{split}\label{marginal_uk_etgp_prior}
\end{equation}
The overall joint can be derived following a similar procedure. Note that we have:
\begin{equation}
\begin{aligned}
        &\fnull{},\unull{} \hspace{-0.4cm}&& \sim p(\fnull{},\unull{}\mid\Xsamples{},\Zsamples{},\lambda)  \\ 
        & \fK{1} \hspace{-0.4cm}&&= \Gcompcinpdep{1}{\Xsamples}(\fnull{});\,\, &&\fK{1} \hspace{-0.4cm}&&= \Gcdirinvcompinpdep{1}{2}{\Xsamples}(\fK{2});\,\, &&\fK{1}  \hspace{-0.4cm}&&= \Gcdirinvcompinpdep{1}{3}{\Xsamples}(\fK{3})\\
         &\uK{1}\hspace{-0.4cm} &&= \Gcompcinpdep{1}{\Zsamples{}}(\unull{});\,\, &&\uK{1} \hspace{-0.4cm}&&= \Gcdirinvcompinpdep{1}{2}{\Zsamples{}}(\uK{2});\,\, &&\uK{1}  \hspace{-0.4cm} \hspace{-0.4cm}&&= \Gcdirinvcompinpdep{1}{3}{\Zsamples{}}(\uK{3})\\ \nonumber
         &\fK{2} \hspace{-0.4cm}&&= \Gcompcinpdep{2}{\Xsamples}(\fnull{});\,\, &&\fK{2} \hspace{-0.4cm}&&= \Gcdirinvcompinpdep{2}{1}{\Xsamples}(\fK{1});\,\, &&\fK{2}  \hspace{-0.4cm}&&= \Gcdirinvcompinpdep{2}{3}{\Xsamples}(\fK{3})\\
         &\uK{2} \hspace{-0.4cm}&&= \Gcompcinpdep{2}{\Zsamples{}}(\unull{});\,\, &&\uK{2} \hspace{-0.4cm}&&= \Gcdirinvcompinpdep{2}{1}{\Zsamples{}}(\uK{1});\,\, &&\uK{2}  \hspace{-0.4cm}&&= \Gcdirinvcompinpdep{2}{3}{\Zsamples{}}(\uK{3})\\
         &\fK{3} \hspace{-0.4cm}&&= \Gcompcinpdep{3}{\Xsamples}(\fnull{});\,\, &&\fK{3} \hspace{-0.4cm} &&= \Gcdirinvcompinpdep{3}{1}{\Xsamples}(\fK{1});\,\, &&\fK{3}  \hspace{-0.4cm}&&= \Gcdirinvcompinpdep{3}{2}{\Xsamples}(\fK{2})\\
         &\uK{3} \hspace{-0.4cm}&&= \Gcompcinpdep{3}{\Zsamples{}}(\unull{});\,\, &&\uK{3} \hspace{-0.4cm}&&= \Gcdirinvcompinpdep{3}{1}{\Zsamples{}}(\uK{1});\,\, &&\uK{3}  \hspace{-0.4cm}&&= \Gcdirinvcompinpdep{3}{2}{\Zsamples{}}(\uK{2})\\
\end{aligned}
\end{equation}
Following similar ideas as before, the pivots are now defined to be $\fK{1}$ and $\uK{1}$. We can also apply a similar conditional independence, and note that the joint distribution over the non \emph{pivots} $\fK{},\uK{}$ also factorizes. Conditional independence holds because any $\fK{c},\uK{c}$ only depends on $\fK{1},\uK{1}$ by a direct mapping; and the conditional distribution over the non \emph{pivots} $p(\fK{c},\uK{c}\mid \fK{1},\uK{1})=p(\fK{c}\mid \fK{1})p(\uK{c}\mid \uK{1})$ factorizes since $\fK{c}$ only depends on $\fK{1}$ and $\uK{c}$ on $\uK{1}$. Writing:
\begin{equation}
    \begin{split}
    & p(\fK{1},\uK{1},\fK{2},\uK{2},\fK{3},\uK{3}) = \\
    & p(\fK{1},\uK{1}) p(\fK{2},\uK{2}\mid\fK{1},\uK{1}) p(\fK{3},\uK{3}\mid\fK{1},\uK{1},\cancel{\fK{2},\uK{2}}) = \\
    & p(\fK{1},\uK{1}) p(\fK{2}\mid\fK{1})p(\uK{2}\mid\uK{1}) p(\fK{3}\mid\fK{1})p(\uK{3}\mid\uK{1}) = \\
        & \underbrace{\underbrace{p(\fnull{}\mid\unull{})\Jacobiancinputdep{\fK{1}}{1}{\Xsamples}}_{p(\fK{1}\mid\uK{1})}\underbrace{p(\unull{})\Jacobiancinputdep{\uK{1}}{1}{\Zsamples{}}}_{p(\uK{1})}}_{p(\fK{1},\uK{1})} \\
    & \underbrace{\mydeltafinpdep{2}{1}{\Xsamples}\mydeltauinpdep{2}{1}{\Zsamples{}}}_{p(\fK{2},\uK{2}\mid\fK{1},\uK{1})=p(\fK{2}\mid\fK{1})p(\uK{2}\mid\uK{1})} \\
    & \underbrace{\mydeltafinpdep{3}{1}{\Xsamples}\mydeltauinpdep{3}{1}{\Zsamples{}}}_{p(\fK{3},\uK{3}\mid\fK{1},\uK{1})=p(\fK{3}\mid\fK{1})p(\uK{3}\mid\uK{1})}
    \end{split}\label{eqn:deriving_pivots_sparse_prior}
\end{equation}
Because we use a diagonal flow, the full Jacobian factorizes as $\Jacobiancinputdep{\fK{1}}{1}{\Xsamples}\Jacobiancinputdep{\uK{1}}{1}{\Zsamples{}}$, allowing us to explicitly write $p(\fK{1}\mid\uK{1})$  and $p(\uK{1})$ (see appendix of \citep{TGP_maronas}). Thus, the overall joint distribution is given by:
\begin{equation}
    \begin{split}
         &  \hspace{-0.75cm}p(\fallk,\uallk,\Wall\mid\Gallcomp,\Xsamples,\Zsamples{},\lambdaall,\nu) = \myprod{c = 1}{C} p(\W{c}\mid\lambdac{c})\\
         & \hspace{-0.75cm}\underbrace{p(\fnull{}\mid \unull{}, \Xsamples,\Zsamples{},\nu) \Jacobiancinputdep{\fpos{k}{1}}{1}{\Xsamples}   \myprod{c = 2}{C}\mydeltafinpdep{c}{1}{\Xsamples{}}}_{p(\fallk\mid\uallk)}\\
         &  \hspace{-0.75cm} \underbrace{p(\unull{}\mid\Zsamples{},\nu)   \Jacobiancinputdep{\upos{k}{1}}{1}{\Zsamples{}}  \myprod{c = 2}{C}\mydeltauinpdep{c}{1}{\Zsamples{}}}_{p(\uallk)}
    \end{split}\label{joint_etgp_sparse_prior}
\end{equation}
where now the \emph{pivots} are $\uK{1}$ and $\fK{1}$. To fully characterize the joint distribution we shall derive where the expression for $p(\fallk\mid\uallk)$ in \ueqn \ref{joint_etgp_sparse_prior} comes from. This conditional distribution is derived by inspection as follows. We factorize the joint distribution in the following two equivalent ways:
\begin{equation}
    \begin{split}
    & p(\fK{1},\fK{2},\fK{3},\uK{1},\uK{2},\uK{3})     = \\
    & p(\fK{1},\fK{2},\fK{3}\mid\uK{1},\uK{2},\uK{3})p(\uK{1},\uK{2},\uK{3}) = \\
    & p(\fK{1},\uK{1}) p(\fK{2},\uK{2}\mid\uK{1},\uK{1}) p(\fK{3},\uK{3}\mid\fK{1},\uK{1}) 
    \end{split}\nonumber
\end{equation}
and used the form of the third line, which is the one we know how to write (\ueqn \ref{eqn:deriving_pivots_sparse_prior}), to derive the expression for the second line, which is the object of our interest.
The expression for the third line has already been written and is given by:
\begin{equation}
\begin{split}
     & p(\fK{1},\fK{2},\fK{3},\uK{1},\uK{2},\uK{3}) =\\
     &  \underbrace{p(\fnull{}\mid\unull{})\textcolor{red}{p(\unull{})}\Jacobiancinputdep{\fK{1}}{1}{\Xsamples}\textcolor{red}{\Jacobiancinputdep{\uK{1}}{1}{\Zsamples{}}}}_{ p(\fK{1},\uK{1})}  \\
     &  \underbrace{ \mydeltafinpdep{2}{1}{\Xsamples} \textcolor{red}{\mydeltauinpdep{2}{1}{\Zsamples{}}}}_{p(\fK{2},\uK{2}\mid\uK{1},\uK{1})} \\
     &  \underbrace{\mydeltafinpdep{3}{1}{\Xsamples}  \textcolor{red}{\mydeltauinpdep{3}{1}{\Zsamples{}}}}_{ p(\fK{3},\uK{3}\mid\fK{1},\uK{1})}  \\
\end{split}\label{sparse_joint_etgp_example}
\end{equation}
Then, since we know the form of $p(\uK{1},\uK{2},\uK{3})$:
\begin{equation}
    \begin{split}
     & p(\uK{1},\uK{2},\uK{3}) = \\
     & p(\unull{})\Jacobiancinputdep{\uK{1}}{1}{\Zsamples{}} \\
     & \mydeltau{2}{1} \mydeltau{3}{1}
    \end{split}
\end{equation}
then, by careful inspection of \ueqn \ref{sparse_joint_etgp_example} we can derive the conditional distribution, which is given by:
\begin{equation}
    \begin{split}
       & p(\fK{1},\fK{2},\fK{3}\mid\uK{1},\uK{2},\uK{3}) = \\
       & p(\fnull{}\mid\unull{})\Jacobiancinputdep{\fK{1}}{1}{\Xsamples} \\
       & \mydeltafinpdep{2}{1}{\Xsamples} \mydeltafinpdep{3}{1}{\Xsamples}
    \end{split}
\end{equation}
where we have just seen (marked in red in \ueqn \ref{sparse_joint_etgp_example}) which elements from the full joint belong to the marginal $p(\uK{1},\uK{2},\uK{3})$, and thus the remaining must belong to the conditional. This gives the prior conditional for $C$ processes:
\begin{equation}
\begin{split}
    p(\fallk\mid\uallk,\Wall,\Xsamples,\Zsamples{},\nu) =& \,\, \underbrace{p(\fnull{}\mid\unull{},\Xsamples,\Zsamples{},\nu)\Jacobiancinputdep{\fK{1}}{1}{\Xsamples}}_{p(\fK{1}\mid\uK{1})}\\ 
    & \myprod{c=2}{C}\mydeltafinpdep{c}{1}{\Xsamples}
    \end{split}\label{conditional_sparse_prior}
\end{equation}
matching the result in \ueqn \ref{joint_etgp_sparse_prior}. With this, we are now ready to derive the sparse variational inference algorithm.

\subsubsection{Marginal variational distribution $q(\fallk)$}

The variational distribution is defined following the ideas from \citep{TGP_maronas,VIinducingpoints_titsias,GPsBigData_hensman}:
\begin{equation}
    q(\fallk,\uallk,\Wall) = p(\fallk\mid\uallk,\Wall) q(\uallk\mid \Wall) q(\Wall)
\end{equation}
where we use the conditional model's prior derived in the previous section and a marginal conditional variational distribution that is defined by warping a multivariate Gaussian in the original \GP{} space $q(\unull{}\mid\varm{},\varS{})$ using $\Gall$, 
where $\varm{} \in \mathds{R}^M,\,\varS{} \in \mathds{R}^{M\times M}$, with the flows from the prior $\Gallcomp$:
\begin{equation}
    \begin{split}
         q(\uallk \mid \varm{},\varS{},\Wall,\Zsamples{}) =\,\, &q(\unull{}\mid \varm{},\varS{}) \Jacobiancinputdep{\upos{k}{1}}{1}{\Zsamples{}} \\
         & \myprod{c= 2}{C}\mydeltauinpdep{c}{1}{\Zsamples{}}
    \end{split}
\end{equation}
and where the distribution over the \NN weights factorizes:
\begin{equation}
    q(\Wall\mid \phiall) = \myprod{c=1}{C}q(\W{c}\mid\phi_c)
\end{equation}
where $\phiall$ denote variational parameters. Note that the dependence of the marginal $q(\uallk)$ on $\Wall$ is required since this distribution is parameterized by the flows of the prior and so inference over $\Wall$ requires dependence between $q(\uallk)$ and $q(\Wall\mid \phiall)$.

To derive our inference algorithm, we need to show how to integrate out inducing points, which turns out that can be done analytically when using diagonal flows, as in \citep{TGP_maronas}:
\begin{equation}
    \begin{split}
    q(\fallk \mid \Wall) =& \int_{\uK{1}} \hdots \int_{\uK{C}} p(\fallk \mid \uallk) q(\uallk)\dd\uK{1}\hdots\dd\uK{C}  \\
    = & \int_{\uK{1}} \hdots \int_{\uK{C}}  p(\fK{1}\mid\uK{1})\myprod{c=2}{C}\mydeltafinpdep{c}{1}{\Xsamples} \\
    & q(\uK{1})\myprod{c=2}{C}\mydeltauinpdep{c}{1}{\Zsamples{}}  \dd\uK{1}\hdots\dd\uK{C}  \\
    = &  \myprod{c=2}{C}\mydeltafinpdep{c}{1}{\Xsamples} \\
    & \int_{\uK{1}} \hdots \int_{\uK{C}} 
    p(\fK{1}\mid\uK{1})q(\uK{1})\myprod{c=2}{C}\mydeltauinpdep{c}{1}{\Zsamples{}} 
    \dd\uK{1}\hdots\dd\uK{C}  \\
    = & \myprod{c=2}{C}\mydeltafinpdep{c}{1}{\Xsamples}  \int_{\uK{1}} p(\fK{1}\mid\uK{1})q(\uK{1}) 
    \dd \uK{1} \\
    = & \myprod{c=2}{C}\mydeltafinpdep{c}{1}{\Xsamples}  \Jacobiancinputdep{\fK{1}}{1}{\Xsamples} \\
    & \int p(\fnull{}\mid\unull{})q(\unull{}) 
    \dd \unull{} \\
    = &  q(\fnull{}) \Jacobiancinputdep{\fK{1}}{1}{\Xsamples} \myprod{c=2}{C}\mydeltafinpdep{c}{1}{\Xsamples} \\
    \end{split}
\end{equation}
where  $q(\fnull{})$ is given by \ueqn \ref{svgp_qf0}. Note that the form of this marginal variational distribution  $q(\fallk{})$ implies the following generative procedure already used in the definition of the \ETGP:
\begin{equation}
\begin{split}
    & \fnull{} \sim q(\fnull{})\\
    \fK{1} = \Gcompcinpdep{1}{\Xsamples}(\fnull{}),\,\,\, & \fK{2} = \Gcompcinpdep{2}{\Xsamples}(\fnull{})\,\,\, \hdots \,\,\, \fK{C} = \Gcompcinpdep{C}{\Xsamples}(\fnull{})
\end{split}
\end{equation}
The sequence of steps used in the derivation are the followings. We start by writing the marginalization in terms of the conditional distribution $p(\fallk\mid\uallk,\Wall)$ and the proposed marginal variational $q(\uallk\mid\Wall)$. From second to third equality we take out from the integral the terms that do not depend on $\uallk$. Then from third to fourth equality we integrate out all $\uallk$ except the \emph{pivot} $\uK{1}$. Note that integration here is straightforward since the Dirac measure integrates to $1$\footnote{Readers concerned with the integration of the Dirac measure in this context can replace it by a Gaussian density taking the limit of $\sigma \rightarrow 0$.}. This let us with one integral over $\uK{1}$. From fourth to fifth equality, we  write $p(\fK{1}\mid\uK{1})$ using the expression in \ueqn \ref{conditional_sparse_prior}, and since the Jacobian does not depend on $\uK{1}$ it is taken out from the integral. Lastly, we apply the \LOTUS rule (see appendix in \citep{TGP_maronas} and below) by noting an expectation over $q(\uK{1})$, which give us a simple Gaussian integral, from which analytical solution $q(\fnull{})$ is well known. This distribution coincides with the \SVGP marginal variational given by \ueqn \ref{svgp_qf0}, as in \TGP{}s.

For self-contained purposes we copy the \LOTUS rule definition in \citep{TGP_maronas}:

\paragraph{\LOTUS rule: } Given an invertible transformation $\G$, and the distribution $p(\fK{})$ induced by transforming samples from a base distribution $p(\fnull{})$, then it holds that expectations of any function $h()$ under $p(\fK{})$ can be computed by integrating \wrt the base distribution $p(\fnull{})$. This is formally known as probability under change of measure. Formally, the above statement implies:
\begin{equation}
    \Expectation{p(\fK{})}{h(\fK{})} = \Expectation{p(\fnull{})}{h(\G(\fnull{}))}
\end{equation}
\subsubsection{Evidence Lower Bound \ELBO}
The Evidence Lower Bound resulting from the prior model and the variational approximate posterior can be written down as:
\begin{equation}
 \begin{split}
     \ELBO & = \int_{\fallk} \int_{\uallk}  \int_{\Wall} q(\fallk,\uallk\mid\Wall)q(\Wall)  \frac{\log p(\Ysamples\mid\fallk) p(\fallk,\uallk\mid\Wall)p(\Wall)}{q(\fallk,\uallk\mid\Wall)q(\Wall)} \dd\fallk\dd\uallk \dd\Wall\\
           & =  \underbrace{\int_{\fallk} \int_{\uallk}  \int_{\Wall} q(\fallk,\uallk\mid\Wall)q(\Wall)   \log p(\Ysamples\mid\fallk)  \dd\fallk\dd\uallk\dd\Wall}_{\ELL}\\
           &   \underbrace{+  \int_{\fallk} \int_{\uallk}  \int_{\Wall} q(\fallk,\uallk\mid\Wall)q(\Wall)   \log\frac{  p(\fallk,\uallk\mid\Wall)p(\Wall)}{q(\fallk,\uallk\mid\Wall)q(\Wall) } \dd\fallk\dd\uallk\dd\Wall}_{-\KLD}
 \end{split}   
\end{equation}
where we again drop the conditioning set, except $\Wall$, for clarity. Working each term separately yields:
\begin{equation}
    \begin{split}
        -\KLD &=  \int_{\fallk} \int_{\uallk}  \int_{\Wall} q(\fallk,\uallk\mid\Wall)q(\Wall)  \log \frac{ p(\fallk,\uallk\mid\Wall)p(\Wall)}{q(\fallk,\uallk\mid\Wall)q(\Wall) } \dd\fallk\dd\uallk\dd\Wall \\
       &= \int_{\fallk} \int_{\uallk} \int_{\Wall} q(\fallk,\uallk\mid\Wall)q(\Wall)  \log\frac{\cancel{p(\fallk\mid\uallk,\Wall)}p(\uallk\mid\Wall)p(\Wall)}{\cancel{p(\fallk\mid\uallk,\Wall)}q(\uallk\mid\Wall)q(\Wall)} \dd\fallk\dd\uallk\dd\Wall  \\
       &=\int_{\fallk} \int_{\uallk} \int_{\Wall} q(\fallk,\uallk\mid\Wall)q(\Wall)  \log\frac{ \cancel{\myprod{c=2}{C}\mydeltauinpdep{c}{1}{\Zsamples{}}}}{\cancel{\myprod{c=2}{C}\mydeltauinpdep{c}{1}{\Zsamples{}}}} \dd\fallk\dd\uallk\dd\Wall  \\
       &+\int_{\fallk} \int_{\uallk} \int_{\Wall} q(\fallk,\uallk\mid\Wall)q(\Wall)  \log\frac{ p(\uK{1}\mid\W{1})p(\Wall)}{q(\uK{1}\mid\W{1})q(\Wall)} \dd\fallk\dd\uallk\dd\Wall  \\
        &= \int_{\uallk} \int_{\Wall} q(\uK{1}\mid\W{1}) \myprod{c=2}{C}\mydeltauinpdep{c}{1}{\Zsamples{}} q(\Wall) \log\frac{p(\uK{1}\mid\W{1})p(\Wall)}{q(\uK{1}\mid\W{1})q(\Wall)} \dd\uallk \dd\Wall \\
        &= \int_{\uK{1}} \int_{\Wall} q(\uK{1}\mid\W{1}) q(\Wall) \log\frac{p(\uK{1}\mid\W{1})p(\Wall)}{q(\uK{1}\mid\W{1})q(\Wall)} \dd\uK{1}\dd\Wall \\
        &= \int_{\Wall}  q(\Wall) \int_{\uK{1}} q(\uK{1}\mid\W{1}) \log\frac{p(\uK{1}\mid\W{1})}{q(\uK{1}\mid\W{1})} \dd\uK{1}\dd\Wall + \int_{\Wall} q(\Wall)\log\frac{p(\Wall)}{q(\Wall)}\dd\Wall \\
        &= \int_{\Wall} q(\Wall)\int_{\unull{}} q(\unull{})  \log\frac{p(\unull{})}{q(\unull{})} \dd\unull{}\dd\Wall + \int_{\Wall} q(\Wall)\log\frac{p(\Wall)}{q(\Wall)}\dd\Wall\\
        &= -\KLD[q(\unull{})||p(\unull{})]-\KLD[q(\Wall{})||p(\Wall{})]
    \end{split}
    \label{KLD_ETGP_complete}
\end{equation}
where in the second and third equalities we cancel common terms. From equality $3$ to $4$ we integrate out all  $\fallk{}$ since nothing depends on them.
In step from equality $4$ to $5$ we integrate out the Dirac measures over all the non-\emph{pivot} elements. From equality $5$ to $6$ we separate expectations and step $6$ to $7$ can be derived in two ways. First, since \KLD is invariant under a parameter transformation (reparameterization) and both the prior and variational distributions are transformed with the same warping function $\Gallcomp$, then the \KLD can be written as that on the original \GP{} space. Another way to derive this \KLD is by noting an expected value of a log-ratio \wrt $q(\uallk)$, allowing us to apply the \LOTUS rule, and corresponding Jacobian cancellations. More precisely:
\begin{equation}
    \begin{split}
        & \int_{\uK{1}} q(\uK{1})  \log\frac{p(\uK{1})}{q(\uK{1})} \dd\uK{1} \\
        &= \int_{\uK{1}} q(\uK{1})  \log\frac{p(\unull{}\mid\Z{},\nu)   \cancel{\Jacobiancinputdep{\upos{k}{1}}{1}{\Zsamples{}} }}{q(\unull{}\mid \varm{},\varS{}) \cancel{\Jacobiancinputdep{\upos{k}{1}}{1}{\Zsamples{}}}} \dd\uK{1} \\
        &= \int_{\unull{}} q(\unull{})  \log\frac{p(\unull{}\mid\Z{},\nu)   }{q(\unull{}\mid \varm{},\varS{})} \dd\unull{} 
     \end{split}
\end{equation}
which is a similar derivation to that in \citep{TGP_maronas}. Note that we could also recognize the \LOTUS rule being applied from equality $3$ to equality $7$ directly in \ueqn \ref{KLD_ETGP_complete}, by previously integrating out $\fallk{}$ and without the $\delta(\cdot)$ cancellations. In other words, we can see the full \KLD over $\uallk{}$ as a direct reparameterization applied to $\unull{}$.

We next derive the \ELL:
\begin{equation}
    \begin{split}
        \ELL &= \int_{\Wall}\int_{\fallk} \int_{\uallk} q(\fallk,\uallk\mid\Wall)q(\Wall)  \log p(\Ysamples\mid\fallk)  \dd\fallk\dd\uallk \dd\Wall\\
        &= \int_{\Wall}\int_{\fallk} q(\fallk\mid\Wall) q(\Wall) \log p(\Ysamples\mid\fallk)  \dd\fallk \dd\Wall\\
        &= \int_{\Wall}\int_{\fnull{}} q(\fnull{}) q(\Wall) \log p(\Ysamples\mid\Gallcompinpdep{\Xsamples}(\fnull{}))  \dd\fnull{}\dd\Wall\\
        &= \int_{\Wall}\int_{\fnull{}} q(\fnull{}) q(\Wall) \log \myprod{n=1}{N}\myprod{c=1}{C}\pi_c(\Gall_{\theta_K(\Wall,\X{n})}(\fnbpos{0,n}{}))^{\mathbb{I}(\Y{n}=c)}  \dd\fnull{}\dd\Wall\\
        &= \mysum{n=1}{N}\mysum{c=1}{C} \mathbb{I}(\Y{n}=c)\int_{\Wall} q(\Wall) \int_{\fnbpos{0,n}{}} q(\fnbpos{0,n}{})  \log \pi_c(\Gall_{\theta_K(\Wall,\X{n})}(\fnbpos{0,n}{})) \dd\fnbpos{0,n}{}\dd\Wall \\
        &\approx \mysum{n=1}{N}\mysum{c=1}{C} \mathbb{I}(\Y{n}=c)\frac{1}{S}\mysum{s=1}{S} \int_{\fnbpos{0,n}{}} q(\fnbpos{0,n}{})  \log \pi_c(\Gall_{\theta_K(\Wall^{s},\X{n})}(\fnbpos{0,n}{})) \dd\fnbpos{0,n}{}; \Wall_s \sim q(\Wall)
    \end{split}
\end{equation}
where we first integrate out $\uallk{}$ yielding the derived conditional marginal $q(\fallk\mid\Wall)$ and then apply the \LOTUS rule to expectation \wrt $q(\fallk{}\mid\Wall{})$. The remaining steps are similar to \SVGP when pluging the specific Categorical Likelihood used in this work.

Using both derivations we recover the \ELBO in the main paper (\ueqn \ref{elbo}):
\begin{equation}
   \begin{split}
    \ELBO =&  \mysum{n=1}{N}\mysum{c=1}{C} \mathbb{I}(\Y{n}=c)\int_{\Wall} q(\Wall) \int_{\fnbpos{0,n}{}} q(\fnbpos{0,n}{})  \log \pi_c(\Gall_{\theta_K(\Wall,\X{n})}(\fnbpos{0,n}{})) \dd\fnbpos{0,n}{}\dd\Wall \\
           &  -\KLD[q(\unull{})||p(\unull{})]-\KLD[q(\Wall{})||p(\Wall{})]
   \end{split} 
   \label{elbo_apx}
\end{equation}

\subsubsection{Computational advantages}

We highlight differences between our proposed model and \SVGP{}s. First, expectations \wrt $q(\fnull{})$ in \ueqn \ref{elbo_apx} can be computed with $1$-d quadrature. By contrast, the \SVGP{} method cannot use quadrature methods and require Monte Carlo. This makes our algorithm computationally advantageous. On the other side expectations \wrt the \NN{}'s parameters can be computed using batched matrix multiplications and in practice we use Monte Carlo Dropout \citep{MCdropout} with one Monte Carlo sample for training, making this computation very efficient.
Moreover, the number of \GP{} operations is constant with $C$. To get  $q(\f{c})$ in \SVGP{}s  one needs a cubic operation to invert $\GPcovar{c}(\Zsamples{},\Zsamples{})$ and $M^2$ operation to compute the variational parameters per class and datapoint, giving a complexity of $\complexity{CM^3 + CNM^2}$. This can be alleviated by 
sharing $K_\nu$ and $\Zsamples{}$ across \GP{}s, resulting in
$\complexity{M^3 + CNM^2}$, at the cost of limiting expressiveness, as shown in the experiment section. \ETGP cost is always $\complexity{M^3 + NM^2}$ (without considering 
the \NN'{}s computations, which for the architecture presented is often much faster and can be done in parallel to \GP{} operations).

\subsection{Proof of proposition 1}
In this section we prove proposition $1$, which we restate for clarity.
\begin{numberedproposition}{1}
    The joint conditional distribution of $C$ non-stationary \GP{}s obtained via a linear flow is given by:
    \begin{align}
         p(\fallk\mid\Wall{})=\Ngauss{\fK{1}\mid\mathbf{b}^1, \A{1} K_\nu(\Xsamples,\Xsamples) \A{1}^\text{T}} \myprod{c= 2}{C} \mydeltafinpdep{c}{1}{\Xsamples}	\,,
    \end{align}
    with $\bb{1}=(b^1(\X{1}),\ldots,b^1(\X{N}))^\text{T}$ 
	and $\A{1}\in \mathds{R}^{N\times N}$ a diagonal matrix with entries 
	$\aaa{1} = (a^1(\X{1}),\hdots,a^1(\X{N}))^\text{T}$. 
	Each marginal $p(\fK{c})$ is Gaussian with mean and covariance 
	given by $\bb{c}$ and $\A{c} K_\nu(\mathbf{X},\mathbf{X})\A{c}^\text{T}$, respectively. 
	The covariance matrix between the pivot $\fK{1}$ and $\fK{c}$, for $c\neq 1$, is given 
	by: $\aaa{1}{(\aaa{c})}^\text{T} \odot K_\nu(\Xsamples{},\Xsamples{})$ with $\odot$ denoting Hadamart product.
\end{numberedproposition}
The proof is divided in two steps. We first derive the marginal distributions $\{p(\fK{c})\}^C_{c=1}$ and then the covariances. For all the proof we will assume that the \emph{pivot} is $\fK{1}$.
First, since a linear flow is used, we can write the flow mapping over a set of samples $\Xsamples{}=(\X{1},\ldots,\X{N})$ in matrix form as:
\begin{equation}
    \fK{1} = \A{1}\fnull{} + \bb{1}
\end{equation}
with $\A{1}\in \mathds{R}^{N\times N}$ being a diagonal matrix with entries $\aaa{1} = (\aanb{}{1}(\X{1}),\hdots,\aanb{}{1}(\X{N}))^\text{T}$ and $\bb{1}=(\bbnb{}{1}(\X{1}),\ldots,\bbnb{}{1}(\X{N}))^\text{T}$. Thus, using the fact that $p(\fnull{}\mid\Xsamples,\nu)=\Ngauss{\fnull{} \mid \veczero , \GPcovar{}(\Xsamples{},\Xsamples{})}$ and the resulting density when applying a linear transformation to a Gaussian density, the marginal distribution over $\fK{1}$ is:
\begin{equation}
    \Ngauss{\fK{1} \mid \bb{1}, \A{1} K_\nu(\Xsamples,\Xsamples) \A{1}^\text{T}} 
\end{equation}
Note that for non-zero mean \GP{} the mean would be given by $\bb{1}+\A{1}\mu_\nu(\Xsamples)$. To derive the marginal distribution for each $c$ we solve the following integral:
\begin{equation}
    \begin{split}
        p(\fK{c}) &= \int_{\fK{1}} p(\fK{1})p(\fK{c}\mid\fK{1}) \dd\fK{1}\\
        & = \int_{\fK{1}}  \Ngauss{\fK{1} \mid \bb{1}, \A{1} K_\nu(\Xsamples,\Xsamples) \A{1}^\text{T}}  \mydeltafinpdep{c}{1}{\Xsamples}	\dd\fK{1}
    \end{split}
\end{equation}
Before solving it note that for any $c$ we have:
\begin{equation}
    \begin{split}
     \fK{c} &= \Gcdirinvcompinpdep{c}{1}{\Xsamples}(\fK{1}) \\
        &= \overbrace{\A{c} \underbrace{\Ainv{1}\left[\fK{1} - \bb{1}\right]}_{\HH}+ \bb{c}}^{\G}
    \end{split}
\end{equation}
yielding:
\begin{equation}
    \begin{split}
        p(\fK{c}) &= \int_{\fK{1}}  \Ngauss{\fK{1} \mid \bb{1}, \A{1} K_\nu(\Xsamples,\Xsamples) \A{1}^\text{T}}  \delta(\fK{c} -  \A{c} \Ainv{1}\left[\fK{1} - \bb{1}\right] - \bb{c}) 	\dd\fK{1} \\
        &= \int_{\fK{1}}  \Ngauss{\fK{1} \mid \bb{1}, \A{1} K_\nu(\Xsamples,\Xsamples) \A{1}^\text{T}}  \delta(\fK{c} -  \A{c} \Ainv{1}\fK{1} + \A{c}\Ainv{1}\bb{1} - \bb{c}) 	\dd\fK{1} \\
    \end{split}
\end{equation}
We then rewrite this last expression to highlight the integral to be solved:
\begin{equation}
    \begin{split}
        &\int_{\fK{1}}  \Ngauss{\fK{1} \mid \varm{}, \varS{}}  \delta(\fK{c} -  \Q\fK{1} - \rrr) \dd\fK{1}
    \end{split}
\end{equation}
with:
\begin{equation}
    \begin{aligned}
        & \varm{} &&= \bb{1} \\
        & \varS{} &&= \A{1} K_\nu(\Xsamples,\Xsamples) \A{1}^\text{T} \\
        & \Q      &&= \A{c} \Ainv{1} \\
        & \rrr    &&= -\A{c}\Ainv{1}\bb{1} + \bb{c}
    \end{aligned}
\end{equation}
The solution of this integral is obtained by the following procedure. First note that if $\Q=\mb{I}$ we recognize a convolution between a Gaussian and a Dirac delta function, easily solved by applying the selection property of Dirac delta functions:
\begin{equation}
\begin{split}
      & \Ngauss{\fK{c}\mid\varm{},\varS{}} \circledast \delta(\fK{c}-\rrr) \coloneqq \int_{-\infty}^{\infty} \Ngauss{\fK{1} \mid \varm{}, \varS{}}  \delta(\fK{c} -\fK{1} - \rrr) \dd\fK{1}\\
      =& \Ngauss{\fK{c} - \rrr \mid\varm{},\varS{}} = \Ngauss{\fK{c} \mid\varm{} + \rrr,\varS{}}  
\end{split}
\end{equation}
where the last step holds by writing the Gaussian density and checking $\fK{c} - \rrr - \varm{} = \fK{c}- (\varm{}+\rrr)$. For $\Q\neq \mb{I}$, we perform an integration by substitution \footnote{After this derivation we found a simpler way to obtain this solution which is given, for completeness, in \uappendix \ref{alternative_derivation_integral}.}, since there is no way we can write the integral as a convolution between two functions. More precisely let $ u = \Q\fK{1} + \rrr$. We have $\fK{1} = \Qinv(u-\rrr)$ and $|\det \nicefrac{\dd u}{\dd\fK{1}} |=|\det \Q|$ which implies the substitution $\dd \fK{1} = |\nicefrac{1}{\det \Q}| \dd u$. Putting all together we have:
\begin{equation}
    \begin{split}
        p(\fK{c})&=\int_{\fK{1}}  \Ngauss{\fK{1} \mid \varm{}, \varS{}}  \delta(\fK{c} -  \Q\fK{1} - \rrr) \dd\fK{1}\\
        &=\int_{\fK{1}}  \Ngauss{\Qinv(u-\rrr) \mid \varm{}, \varS{}}  \delta(\fK{c} -  u) \left|\frac{1}{\det \Q}\right|\dd u\\
        &= \frac{1}{\left|\det \Q\right|} \Ngauss{\Qinv(\fK{c}-\rrr) \mid \varm{}, \varS{}}
    \end{split}
\end{equation}
Beyond the substitution the integral is solved by applying the selection property of the delta function. After applying some standard algebra to the Gaussian distribution (see \uappendix \ref{algebra_manipulation}) we have:
\begin{equation}
     \Ngauss{\Qinv(\fK{c}-\rrr) \mid \varm{}, \varS{}} = |\det\Q|\,\,\Ngauss{ \fK{c} \mid \Q\varm{} + \rrr ,\Q\varS{}\Q^\text{T}}
\end{equation}
Giving the final result:
\begin{equation}
    p(\fK{c})= \Ngauss{ \fK{c} \mid \Q\varm{} + \rrr ,\Q\varS{}\Q^\text{T}}
    \label{eqn:result_of_integration}
\end{equation}
since $|\det \Q|$ cancels with $\nicefrac{1}{\left|\det \Q\right|}$. Note this result matches the one obtained with the convolution for $\Q=\mb{I}$.
If we now substitute the shortcuts for $\Q,\varS{},\varm{},\rrr$ we have for the covariance:
\begin{equation}
    \begin{split}
       \Q\varS{}\Q^\text{T} &=   \A{c} \Ainv{1}\A{1} K_\nu(\Xsamples,\Xsamples) \A{1}^\text{T}\left[\A{c} \Ainv{1}\right]^\text{T} \\
       &=\A{c} \Ainv{1}\A{1} K_\nu(\Xsamples,\Xsamples) \A{1} \Ainv{1}\A{c}  \\
       &= \A{c} K_\nu(\Xsamples,\Xsamples) \A{c}  
    \end{split}
\end{equation}
where we apply some standard matrix identities. In particular the transpose of the product is the product of the transposes in reverse order, and the transpose of a diagonal matrix is equal to the diagonal matrix. 
For the mean we have:
\begin{equation}
    \begin{split}
        \rrr + \Q\varm{} &= \cancel{-\A{c}\Ainv{1}\bb{1}} + \bb{c} \cancel{+\A{c}\Ainv{1}\bb{1}} = \bb{c}
    \end{split}\label{proposition_proof_mean}
\end{equation}
finishing the first part of the proof, which we re-emphasize:

\emph{The marginal distribution for any $\fK{c}$ has density given by: $\Ngauss{\fK{c} \mid \bb{c},\A{c} K_\nu(\mathbf{X},\mathbf{X})\A{c}^\text{T}}$ }

Importantly, note that if non-zero \GP{}s are used, this result is also matched with the particularity that the mean is given by $\bb{c}+\A{c}\mu_\nu(\Xsamples)$. This is seen by replacing $\A{c}\Ainv{1}\bb{1}$ for $\A{c}\Ainv{1}[\bb{1}+\A{1}\mu_\nu(\Xsamples{})]$ in \ueqn \ref{proposition_proof_mean}.

The second part of the proof characterizes some of the linear dependencies (covariance) in the joint distribution $p(\fallk{})$. We note that in this paper we have not figured out the form of the joint distribution if the \emph{pivot} is integrated out, \ie how the pivot couples the rest of latent functions. In other words, the distribution $p(\fK{2},\hdots,\fK{C}) = \int_{\fK{1}} p(\fK{2},\hdots,\fK{C}\mid\fK{1})p(\fK{1})\dd \fK{1}$ is unknown, which limits the full characterization of the covariances in the joint distribution.

For this reason in this proposition we just characterize the covariances between any $\fK{c}$ and the \emph{pivot} $\fK{1}$. For this we use the expression:
\begin{equation}
    \mathds{COV}[\fK{1},\fK{c}] = \mathds{E}[\fK{1}{(\fK{c})}^\text{T}] - \mathds{E}[\fK{1}]\mathds{E}[\fK{c}]^\text{T}
\end{equation}
The expected values can be directly obtained from the marginal distributions obtained in the first part of the proof. In particular:
\begin{equation}
    \begin{split}
       \Expectation{}{\fK{1}} &= \bb{1} \\
       \Expectation{}{\fK{c}} &= \bb{c} \\
    \end{split}
\end{equation}
To derive the covariance, it is easier to do it by just looking at its entries at two single points locations  $\X{n}$ and $\X{n'}$ and then generalizing the result. Following the main paper notation we have $\fnbpos{0,n}{} \coloneqq \fnbnull{}(\X{n})$, $\bbnb{}{c}(\X{n}) \coloneqq \bbnb{n}{c}$ and $\aanb{}{c}(\X{n}) \coloneqq \aanb{n}{c}$. 
\subsubsection{Covariance between $\fnbpos{K,n}{1}$ and $\fnbpos{K,n}{c}$ at a single location $n$}
For this we compute:
\begin{equation}
    \begin{split}
        & \Expectation{}{\fnbpos{K,n}{1}\fnbpos{K,n}{c}} = \\
        & \int_{\fnbpos{K,n}{1}} \int_{\fnbpos{K,n}{c}} \fnbpos{K,n}{1}\fnbpos{K,n}{c} \Ngauss{\fnbpos{K,n}{1} \mid \bbnb{n}{1},\aanb{n}{1}\GPcovar{}(\X{n},\X{n})\aanb{n}{1}} \delta\left( \fnbpos{K,n}{c} - \frac{\aanb{n}{c}}{\aanb{n}{1}} \left[\fnbpos{K,n}{1}-\bbnb{n}{1}\right]-\bbnb{n}{c} \right) \dd\fnbpos{K,n}{1}\dd\fnbpos{K,n}{c}\\
        &=\int_{\fnbpos{K,n}{1}}\left[ 
          \frac{\aanb{n}{c}}{\aanb{n}{1}}  \fnbpos{K,n}{1}\fnbpos{K,n}{1}
        -\frac{\aanb{n}{c}}{\aanb{n}{1}} \bbnb{n}{1} \fnbpos{K,n}{1}
        + \bbnb{n}{c}\fnbpos{K,n}{1} 
        \right]
        \Ngauss{\fnbpos{K,n}{1} \mid \bbnb{n}{1},\aanb{n}{1}\GPcovar{}(\X{n},\X{n})\aanb{n}{1}} \dd\fnbpos{K,n}{1}\\
        &=  
        \frac{\aanb{n}{c}}{\aanb{n}{1}}\left[\aanb{n}{1}\GPcovar{}(\X{n},\X{n})\aanb{n}{1} + {\left(\bbnb{n}{1}\right)}^2 \right] 
        -\frac{\aanb{n}{c}}{\aanb{n}{1}} {\left(\bbnb{n}{1}\right)}^2 
        + \bbnb{n}{c}\bbnb{n}{1} \\
        &= \aanb{n}{c}\GPcovar{}(\X{n},\X{n})\aanb{n}{1} + \bbnb{n}{c}\bbnb{n}{1}
    \end{split}
\end{equation}
yielding,
\begin{equation}
 \begin{split}
     & \mathds{COV}[\fnbpos{K,n}{1},\fnbpos{K,n}{c}] =  \Expectation{}{\fnbpos{K,n}{1}\fnbpos{K,n}{c}} -  \Expectation{}{\fnbpos{K,n}{1}}\Expectation{}{\fnbpos{K,n}{c}} \\
     & = \aanb{n}{c}\GPcovar{}(\X{n},\X{n})\aanb{n}{1} + \bbnb{n}{c}\bbnb{n}{1} -  \bbnb{n}{c}\bbnb{n}{1} \\
     &= \aanb{n}{c}\GPcovar{}(\X{n},\X{n})\aanb{n}{1} 
 \end{split}   
\end{equation}
So the covariance between two processes at all locations $N$ will be given by the diagonal of $\GPcovar{}{}(\Xsamples,\Xsamples)$ element-wise multiplied by the diagonal of $\aaa{1}{(\aaa{c})}^\text{T}$.
\subsubsection{Covariance between $\fnbpos{K,n}{1}$ and $\fnbpos{K,n'}{c}$ at different locations $n,n'$}
We again emphasize the main paper notation for easier presentation and re-highlight that for this particular section $\fK{1}=(\fnbpos{K,n}{1},\fnbpos{K,n'}{1})^\text{T}$ and similar for any parameter short cut  \eg $\bb{1}$. For this we compute:
\begin{equation}
    \begin{split}
        & \Expectation{}{\fnbpos{K,n}{1}\fnbpos{K,n'}{c}} = \\
        & \int_{\fnbpos{K,n}{1}}\int_{\fnbpos{K,n'}{1}} \int_{\fnbpos{K,n'}{c}} \fnbpos{K,n}{1}\fnbpos{K,n'}{c} \Ngauss{\fK{1} \mid \bb{1}, \A{1} K_\nu(\Xsamples,\Xsamples) \A{1}^\text{T}} 
        \delta\left( \fnbpos{K,n'}{c} - \frac{\aanb{n'}{c}}{\aanb{n'}{1}} \left[\fnbpos{K,n'}{1}-\bbnb{n'}{1}\right]-\bbnb{n'}{c} \right) \dd\fnbpos{K,n}{1}\dd\fnbpos{K,n'}{1}\dd\fnbpos{K,n'}{c}\\
        &=\int_{\fnbpos{K,n}{1}}\int_{\fnbpos{K,n'}{1}}\left[ 
         \frac{\aanb{n'}{c}}{\aanb{n'}{1}}\fnbpos{K,n}{1}\fnbpos{K,n'}{1}
         -\frac{\aanb{n'}{c}}{\aanb{n'}{1}}\bbnb{n'}{1}\fnbpos{K,n}{1}
         +\bbnb{n'}{c}\fnbpos{K,n}{1}
        \right]
        \Ngauss{\fK{1} \mid \bb{1}, \A{1} K_\nu(\Xsamples,\Xsamples) \A{1}^\text{T}}\dd\fnbpos{K,n}{1}\dd\fnbpos{K,n'}{1}\\
        &= \frac{\aanb{n'}{c}}{\aanb{n'}{1}}\left[\aanb{n}{1} \GPcovar{}(\X{n},\X{n'})\aanb{n'}{1} + \bbnb{n'}{1}\bbnb{n}{1}\right]
        -\frac{\aanb{n'}{c}}{\aanb{n'}{1}}\bbnb{n'}{1}\bbnb{n}{1}
        +\bbnb{n'}{c}\bbnb{n}{1}\\
        &=\aanb{n'}{c}\aanb{n}{1}\GPcovar{}(\X{n},\X{n'}) +\bbnb{n'}{c}\bbnb{n}{1}
    \end{split}
\end{equation}
yielding,
\begin{equation}
 \begin{split}
     & \mathds{COV}[\fnbpos{K,n}{1},\fnbpos{K,n'}{c}] =  \Expectation{}{\fnbpos{K,n}{1}\fnbpos{K,n'}{c}} -  \Expectation{}{\fnbpos{K,n}{1}}\Expectation{}{\fnbpos{K,n'}{c}} \\
     & = \aanb{n'}{c}\aanb{n}{1}\GPcovar{}(\X{n},\X{n'}) +\bbnb{n'}{c}\bbnb{n}{1} -  \bbnb{n}{1}\bbnb{n'}{c} \\
     &=  \aanb{n'}{c}\aanb{n}{1}\GPcovar{}(\X{n},\X{n'})
 \end{split}   
\end{equation}
Note that from the above integral is easy to see that
$\mathds{COV}[\fnbpos{K,n'}{1},\fnbpos{K,n}{c}] = \aanb{n}{c}\aanb{n'}{1}\GPcovar{}(\X{n'},\X{n})$ .

\subsubsection{Covariance between $\fK{1}$ and $\fK{c}$}

We have derived the covariance between the \emph{pivot} and any other process at the same location $\X{n}$ and between two different locations $\X{n},\X{n'}$. Note that for $N$ arbitrary locations $\Xsamples$, the covariance between any pair of elements can be obtained by any of the two derivations shown above.

In summary, at a given location $\X{n}$ the covariance between two processes is $\aanb{}{c}(\X{n})\aanb{}{1}(\X{n})\GPcovar{}(\X{n},\X{n})$ and at two different locations $\X{n}$ and $\X{n'}$ the covariance between two processes is  $\aanb{}{c}(\X{n'})\aanb{}{1}(\X{n})\GPcovar{}(\X{n},\X{n'})$.

Thus, the covariance abetween the processes at $N$ locations $\Xsamples$ is given by $\aaa{1}{(\aaa{c})}^\text{T} \odot K_\nu(\Xsamples{},\Xsamples{})$. 

\subsection{Alternative derivation of \ueqn \ref{eqn:result_of_integration}}
\label{alternative_derivation_integral}

We found a simpler way of obtaining \ueqn \ref{eqn:result_of_integration} by approximating the Dirac measure with a Gaussian and taking the limit of the variance to $0$. With this, we can apply standard Gaussian integration and yield the desired result.

First:
\begin{equation}
     \delta(\fK{c} -  \Q\fK{1} - \rrr) = \lim_{\lambda \rightarrow 0}\Ngauss{\fK{c}\mid \Q\fK{1} + \rrr, \lambda \mb{I} }
\end{equation}
and so the integral is solved by:
\begin{equation}
    \begin{split}
        p(\fK{c})&= \lim_{\lambda \rightarrow 0} \int_{\fK{1}}  \Ngauss{\fK{1} \mid \varm{}, \varS{}} \Ngauss{\fK{c}\mid \Q\fK{1} + \rrr, \lambda \mb{I} }\dd\fK{1}\\
        &= \lim_{\lambda \rightarrow 0}  \Ngauss{\fK{c}\mid \Q\varm{} + \rrr, \lambda \mb{I} + \Q\varS{}\Q^\text{T} }\\
        &=  \Ngauss{\fK{c}\mid \Q\varm{} + \rrr, \Q\varS{}\Q^\text{T} }
    \end{split}
\end{equation}
\subsection{Algebra manipulation of the Gaussian distribution}
\label{algebra_manipulation}

We are interested in showing:
\begin{equation}
      \Ngauss{\Qinv(\fK{}-\rrr) \mid \varm{}, \varS{}} = \det\Q\,\, \Ngauss{\fK{} \mid \Q\varm{}+\rrr, \Q\varS{}\Q^\text{T}} 
\end{equation}
We provide the steps to be performed since we haven't found the steps available in the references searched. We can show this equivalence either by the technique of completing the square or by making simple manipulations to the exponent in the Gaussian distribution. We assume these Gaussian distributions have dimensionality $n$. Our manipulations use standard matrix operations that can be found in the matrix cookbook \citep{matrixcookbook}.
\subsubsection{Manipulation of the exponent}
We have:
\begin{equation}
    \begin{split}
        &\Ngauss{\Qinv(\fK{}-\rrr) \mid \varm{}, \varS{}} \\ 
        &=\frac{1}{(2\pi)^{\nicefrac{n}{2}} (\det\varS{})^{\nicefrac{1}{2}}}\exp\left\{(\Qinv(\fK{}-\rrr)-\varm{})^\text{T}\Sinv(\Qinv(\fK{}-\rrr)-\varm{})\right\}\\
        &=\frac{1}{(2\pi)^{\nicefrac{n}{2}} (\det\varS{})^{\nicefrac{1}{2}}}\exp\left\{((\Qinv\fK{})^\text{T}-(\Qinv\rrr)^\text{T}-\varm{}{^\text{T}})\Q^\text{T}(\Q^\text{T})^{-1}\Sinv\Qinv\Q(\Qinv\fK{}-\Qinv\rrr-\varm{})\right\}\\
        &=\frac{1}{(2\pi)^{\nicefrac{n}{2}} (\det\varS{})^{\nicefrac{1}{2}}}\exp\left\{((\Qinv\fK{})^\text{T}\Q^\text{T}-(\Qinv\rrr)^\text{T}\Q^\text{T}-\varm{}{^\text{T}}\Q^\text{T})(\Qinv)^\text{T}\Sinv\Qinv(\fK{}-\rrr-\Q\varm{})\right\}\\
        &=\frac{1}{(2\pi)^{\nicefrac{n}{2}} (\det\varS{})^{\nicefrac{1}{2}}}\exp\left\{(\fK{}{^\text{T}}(\Q^\text{T})^{-1}\Q^\text{T}-\rrr^\text{T}(\Q^\text{T})^{-1}\Q^\text{T}-(\Q\varm{})^\text{T})(\Qinv)^\text{T}\Sinv\Qinv(\fK{}-\rrr-\Q\varm{})\right\}\\
        &=\frac{1}{(2\pi)^{\nicefrac{n}{2}} (\det\varS{})^{\nicefrac{1}{2}}}\exp\left\{(\fK{}{^\text{T}}-\rrr^\text{T}-(\Q\varm{})^\text{T})(\Q\varS{}\Q^\text{T})^{-1}(\fK{}-\rrr-\Q\varm{})\right\}\\
        &=\frac{1}{(2\pi)^{\nicefrac{n}{2}} (\det\varS{})^{\nicefrac{1}{2}}}\exp\left\{(\fK{}-\rrr-\Q\varm{})^\text{T}(\Q\varS{}\Q^\text{T})^{-1}(\fK{}-\rrr-\Q\varm{})\right\}\\
    \end{split}
\end{equation}
This gives an unnormalized Gaussian distribution with mean $\Q\varm{}+\rrr$ and covariance $\Q\varS{}\Q^\text{T}$
\subsubsection{Completing the square method}

We can obtain a similar result by the technique of completing the square. Since we know that a scale and shift on a function argument does not change the function shape, \ie scaling and shifting a Gaussian will give a Gaussian curve, we can use the technique of completing the square to recognize the mean and covariance matrix \citep{bishopbook}. %

From:
\begin{equation}
    \begin{split}
 &\Ngauss{\Qinv(\fK{}-\rrr) \mid \varm{}, \varS{}} \\ 
        &=\frac{1}{(2\pi)^{\nicefrac{n}{2}} (\det\varS{})^{\nicefrac{1}{2}}}\exp\left\{(\Qinv(\fK{}-\rrr)-\varm{})^\text{T}\Sinv(\Qinv(\fK{}-\rrr)-\varm{})\right\}\\
    \end{split}
\end{equation}
We expand the quadratic form:
\begin{equation}
    \begin{split}
    & (\Qinv(\fK{}-\rrr)-\varm{})^\text{T}\Sinv(\Qinv(\fK{}-\rrr)-\varm{}) \\
    &= (\Qinv(\fK{}-\rrr))^\text{T}\Sinv(\Qinv(\fK{}-\rrr))
    -2(\Qinv(\fK{}-\rrr))^\text{T}\Sinv\varm{}
    +\varm{}{^\text{T}}\Sinv\varm{} \\
    &= (\Qinv\fK{})^\text{T}\Sinv\Qinv\fK{}
     - (\Qinv\fK{})^\text{T}\Sinv\Qinv\rrr{} 
     - (\Qinv\rrr)^\text{T}\Sinv\Qinv\fK{} \\ 
    & +  (\Qinv\rrr)^\text{T}\Sinv\Qinv\rrr 
      - 2 (\Qinv\fK{})^\text{T}\Sinv\varm{}
      + 2 (\Qinv\rrr{})^\text{T}\Sinv\varm{}
      +\varm{}{^\text{T}}\Sinv\varm{}
    \end{split}
\end{equation}
and first recognize the terms depending quadratically on $\fK{}$:
\begin{equation}
    \begin{split}
          (\Qinv\fK{})^\text{T}\Sinv\Qinv\fK{} &=\fK{}{^\text{T}}(\Qinv)^\text{T}\Sinv\Qinv\fK{}\\
         &= \fK{}{^\text{T}}(\Q^\text{T})^{-1}\Sinv\Qinv\fK{}\\
         &= \fK{}{^\text{T}}(\Q\varS{}\Q^\text{T})^{-1}\fK{}\\
    \end{split}
\end{equation}
Recognizing the covariance to be $\Q\varS{}\Q^\text{T}$. Now looking at the terms that depend linearly on $\fK{}$ we have:
\begin{equation}
    \begin{split}
        & - (\Qinv\fK{})^\text{T}\Sinv\Qinv\rrr{} 
        - (\Qinv\rrr)^\text{T}\Sinv\Qinv\fK{}
        - 2 (\Qinv\fK{})^\text{T}\Sinv\varm{} \\
        &= - \fK{}{^\text{T}}(\Qinv)^\text{T}\Sinv\Qinv\rrr{} 
          - \rrr^\text{T}(\Qinv)^\text{T}\Sinv\Qinv\fK{}
          - 2\fK{}{^\text{T}}(\Qinv)^\text{T}\Sinv\varm{} \\
        &= - \fK{}{^\text{T}}(\Q\varS{}\Q^\text{T})^{-1}\rrr{} 
           - \rrr^\text{T}(\Q\varS{}\Q^\text{T})^{-1}\fK{}
          - 2\fK{}{^\text{T}}(\Qinv)^\text{T}\Sinv\varm{} \\
    \end{split}
\end{equation}
Looking closely at $\rrr^\text{T}(\Q\varS{}\Q^\text{T})^{-1}\fK{}$ we have:
\begin{equation}
    \begin{split}
    \rrr^\text{T}(\Q\varS{}\Q^\text{T})^{-1}\fK{} & =\fK{}{^\text{T}}((\Q\varS{}\Q^\text{T})^{-1})^\text{T}\rrr\\
    &= \fK{}{^\text{T}}((\Q\varS{}\Q^\text{T})^\text{T})^{-1}\rrr\\
    &= \fK{}{^\text{T}}(\Q\varS{}{^\text{T}}\Q^\text{T})^{-1}\rrr\\
    &= \fK{}{^\text{T}}(\Q\varS{}\Q^\text{T})^{-1}\rrr
    \end{split}
\end{equation}
since $\varS{}$ is symmetric \ie $\varS{}=\varS{}{^\text{T}}$. This yields a final linear term given by:
\begin{equation}
   \begin{split}
       & - \fK{}{^\text{T}}(\Q\varS{}\Q^\text{T})^{-1}\rrr{} 
       - \fK{}{^\text{T}}(\Q\varS{}\Q^\text{T})^{-1}\rrr
       - 2\fK{}{^\text{T}}(\Qinv)^\text{T}\Sinv\varm{} \\
   \end{split} 
\end{equation}
By rewriting $- 2\fK{}{^\text{T}}(\Qinv)^\text{T}\Sinv\varm{} $ as:
\begin{equation}
 \begin{split}
    & - 2\fK{}{^\text{T}}(\Qinv)^\text{T}\Sinv\varm{} = - 2\fK{}{^\text{T}}(\Qinv)^\text{T}\Sinv\Qinv\Q\varm{}  \\
    &= - 2\fK{}{^\text{T}}(\Q\varS{}\Q^\text{T})^{-1}\Q\varm{}
 \end{split}   
\end{equation}
We obtain the final desired linear term:
\begin{equation}
\begin{split}
    & - \fK{}{^\text{T}}(\Q\varS{}\Q^\text{T})^{-1}\rrr 
       - \fK{}{^\text{T}}(\Q\varS{}\Q^\text{T})^{-1}\rrr
       - 2\fK{}{^\text{T}}(\Qinv)^\text{T}\Sinv\varm{} =\\
     & -2 \fK{}{^\text{T}}(\Q\varS{}\Q^\text{T})^{-1}\rrr
       -2 \fK{}{^\text{T}}(\Q\varS{}\Q^\text{T})^{-1}\Q\varm{} \\
     &=-2 \fK{}{^\text{T}}(\Q\varS{}\Q^\text{T})^{-1}[\Q\varm{}+\rrr]
\end{split}
\end{equation}
Recognizing $\Q\varm{}+\rrr$ as the mean of the distribution. 

Finally, we would have to check if there is a way for the remaining constant terms to be re-written as $[\Q\varm{}+\rrr]^\text{T}\Sinv[\Q\varm{}+\rrr]$ otherwise the probability density would not be correctly normalized. Since we know from the previous section that this yields an unnormalized density we omit this step. Anyway we can skip this step since we know that function argument scaling is a non-volume preserving transformation, something that can be trivially checked by computing the area of a function $x(t) = u(t)-u(t-1)$  and its scaled version $x(2t)$, where $u(t)$ is the unit (or Heaviside) step function.

This can be more formally check if we consider the probability under change of variable. If we have $\Ysamples{}=f(\X{})$ for some invertible function $f(\cdot)$ with $\Ysamples{}\sim p(\Ysamples{})$, then:

\begin{equation}
    p(\X{}) = p(\Ysamples{}=f(\X{})) \left|\det \frac{\nabla f(\X{})}{\X{}} \right|
\end{equation}
for a linear transformation $\Ysamples{}=\Q\X{} + \rrr$, this gives:
\begin{equation}
      p(\X{}) = p(\Ysamples{}=\Q\X{} + \rrr) \left|\det \Q \right|
\end{equation}
formally showing that a scale in the argument of a density implies a non-volume preserving transformation and thus without the Jacobian correction it would not be a proper normalized density.

\subsubsection{The remaining normalization constant}

It is clear that the scaling of the Gaussian argument gives an unnormalized density with mean $\Q\varm{}+\rrr$ and covariance $\Q\varS{}\Q^\text{T}$. A proper normalized Gaussian density would have a multiplication constant equal to:
\begin{equation}
    \frac{1}{(2\pi)^{\nicefrac{n}{2}} (\det\Q\varS{}\Q^\text{T})^{\nicefrac{1}{2}}}
\end{equation}
but our result has:
\begin{equation}
    \frac{1}{(2\pi)^{\nicefrac{n}{2}} (\det\varS{})^{\nicefrac{1}{2}}}
\end{equation}
Operating the determinant we see:
\begin{equation}
    \begin{split}
        & (\det\Q\varS{}\Q^\text{T})^{\nicefrac{1}{2}} = (\det\Q \det\varS{} \det \Q^\text{T})^{\nicefrac{1}{2}} \\
        & =(\det\varS{})^{\nicefrac{1}{2}} (\det\Q \det \Q)^{\nicefrac{1}{2}} = (\det\varS{})^{\nicefrac{1}{2}} |\det\Q|  
    \end{split}
\end{equation}
where since the determinant is a scalar value we know that $\sqrt{x^2}=|x|$. This means that we need to scale our density by $\nicefrac{1}{|\det \Q|}$, or, in other words, our density has been unnormalized by multiplying it by $|\det \Q|$. With this, we conclude:
\begin{equation}
      \Ngauss{\Qinv(\fK{}-\rrr) \mid \varm{}, \varS{}} = |\det\Q|\,\, \Ngauss{\fK{} \mid \Q\varm{}+\rrr, \Q\varS{}\Q^\text{T}} 
\end{equation}

}
\fi

%% file: sections/Z2_experiments_appendix.tex
\section{Experiment Appendix}
\label{experiments_appendix}

\ifapendixsubmission
{This information is attached in the supplementary  material.}
\else
{

In this appendix we provide training/evaluation details alongside additional results.

Information about the different datasets used can be found in the code, where a link to the website of each particular dataset can be found. Information about dataset preprocessing can also be found in the code since datasets are very different (images, tabular, tabular with discrete inputs \etc) and so different preprocessing was done. General preprocessing steps are the normalization to $[0,1]$ range of image datasets, normalization by the mean and standard deviation in continuous tabular datasets, and different normalization procedures depending on the type of feature. For example, a dataset containing working age information was normalized by dividing by $65$ since that is the maximum number of years (on average) of a standard worker's life. 

Common to all experiments is the following information. Experiments are run using \GPFLOW \citep{GPflow2017,GPflow2020multioutput}. Unless mentioned we use default \GPFLOW parameters.  Inducing points are initialized using Kmeans algorithm for \texttt{vowel},\texttt{absenteeism} and \texttt{avila} with $10$ reinitializations and parallel Kmeans for \texttt{characterfont} and \texttt{devangari} with $3$ reinitializations. The length scale of \RBF kernels was initialized to $2.0$ and the mixing matrix randomly. Non-stationary kernels are initialized with a length scale of $2.0$ for the arcosine and with an identity matrix for the Neural Network kernel. All kernels employ automatic relevance determination if possible. The variational mean is initialized to zero and the Cholesky factorization of the variational covariance to the identity matrix multiplied by $1e-5$. In all the experiments the model used to compute the train/valid/test metrics was the model corresponding to the epoch with best (highest) \ELBO. We use Adam optimizer.

For all the \SVGP models we run models with learning rate values of $0.01$ and $0.001$. For certain choices of hyper-parameters if we saw that $0.01$ was providing better results than $0.001$ we keep searching just with $0.01$. In some cases we also look for other learning rates \eg $0.05$ in light of finding the best baseline model to compare against. We run either $10000$ or $15000$ epochs for \texttt{vowel},\texttt{absenteeism} and \texttt{avila} and $100,200,500,1000,2000$ epochs for \texttt{characterfont} and \texttt{devangari}. For these last two dataset we do not always launch $2000$ epochs, and only did it if we found a big increase in performance from the run with $500$ to $1000$ epochs. Note that training times are average over epochs and we do not provide the full time of the experiment (which in turns imply that the \ETGP is even faster since we run them just for $500$ epochs). We run models with number of inducing points $\{100,50,20\}$  for (\texttt{vowel},\texttt{absenteeism} and \texttt{avila}) and $100$ for \texttt{characterfont} and \texttt{devangari}. We also experiment with the parameters of the covariance (including the mixing matrix parameters in \RBFCORR) being freezed for $2000$ (\texttt{vowel},\texttt{absenteeism} and \texttt{avila}) or $50$ (\texttt{characterfont} and \texttt{devangari}) epochs or trained end to end, \ie no freezing is applied, following \citep{TGP_maronas,GPclassificationHensman}. Once all these experiments were launched, we select for each set of kernel, number of inducing points etc, the model giving the best performance by directly looking at the test set, in order to evaluate the proposed model in the most optimistic situation for each \SVGP{} baseline.

For the \ETGP model selection was done using a validation split with different number of points per dataset. This information is provided by looking at the code that loads the data. For the \ETGP all the models are run for $15000$ epochs for \texttt{vowel},\texttt{absenteeism} and \texttt{avila}  and $500$ epochs for \texttt{characterfont} and \texttt{devangari} (which implies that the total training time of our models is even faster), and the best selected model on validation for $100$ inducing points, is run for $50$ and $20$, in contrast with \SVGP where each $50$ and $20$ inducing points model can have its own set of training hyperparameters. Bayesian flows are trained with $1$ Monte Carlo dropout sample and evaluated (\ie posterior predictive computation) using $20$ dropout samples. The learning rate experimented was $0.01$ and $0.001$ and all the parameters are trained from the beginning without freezing. The \NN architectures were chosen depending on the input size of the dataset. All these architectures have an input layer equal to the dimensionality of the data and an output layer given by the number of parameters of the flow multiplied by the number of classes. We tested \LINEAR, \SAL\citep{compositionally_warped_gp_rios:2019} with length $3$ and \TANH \citep{WarpedGPs} with length $3$ and $4$ elements in the linear combination. The length of the flow corresponds to the value of $K$ in the flow parameterization, \ie it is the number of, \eg individual \SAL transformations, being concatenated. All the \NN use hyperbolic tangent activation function and we use a variance of a Gaussian prior over flow parameters set to $5000,50000,50000$ which corresponds to a weight decay factor of $1e-4,1e-5,1e-6$ without considering the constant value of the Gaussian prior that depends on the number of parameters. For \texttt{vowel},\texttt{absenteeism} and \texttt{avila} we test networks with $0,1,2$ hidden layers with $25,50,100$ neurons per layer and with dropout probabilities of $0.25,0.5,0.75$ except \texttt{avila}  that only uses $0.25,0.5$. We tested $0.75$ to see if higher uncertainty in the \NN posterior could help in regularizing the datasets with fewer number of training points. For \texttt{devangari} we test $0,1,2$ hidden layers with $512,1024$ neurons per layer. We also tested a projection network of $0,1$ hidden layers with $512$ neurons per hidden layer and $256$ neurons per output layer. The output of this projection network is feed into another neural network that maps the $256$ dimensions to the number of parameters. This second \NN  has $0,1$ hidden layers with $256,128$ neurons per layer. All these networks have a dropout probability of $0.5$. For \texttt{characterfont} we also use a dropout probability of $0.5$ and \NN with $0,1,2$ hidden layers with $256$ neuron per layer. We also test projection networks of $0,1,2$ hidden layers with $512,256$ neuron per hidden layer and output layer of $256$ neurons. This is then feed into another neural network with $0,1,2$ hidden layers and $256$ neuron per layer.

Regarding the initialization of the flows we follow \citep{TGP_maronas} and initialize the flows to the identity by first learning the identity mapping using a non-input dependent flow, and then learning the parameters of the neural network to match each point in the training dataset to the learned non-input dependent parameters. Both initialization procedures are launched $5$ times with a learning rate of $0.05$ and Adam optimizer for any dataset and flow architecture. The input dependent initialization is run for $1000$ epochs in \texttt{vowel},\texttt{absenteeism} and \texttt{avila} and for $100$ epochs in \texttt{characterfont} and \texttt{devangari}. Some preliminary runs were done to test if these hyperparameters allow the flow to be properly initialized and then all these parameters were used for any flow initialization in our validation search without further analysis. We found in general that with fewer epochs the flow could be also initialized properly, but decided to run a considerable number of initialization epochs. We highlight that this procedure can be done in parallel to Kmeans initialization, for readers concerned with the training time associated with this initialization procedure.

\subsection{Log Likelihood results}
In this subsection we provide the corresponding \LL results in \fig \ref{fig:log-likelihood_stationary_mixing} and \fig \ref{fig:log-likelihood_non_stationary} which are discussed in the main text.
\begin{figure}[!htb]
    \centering
    \includegraphics[width=\textwidth]{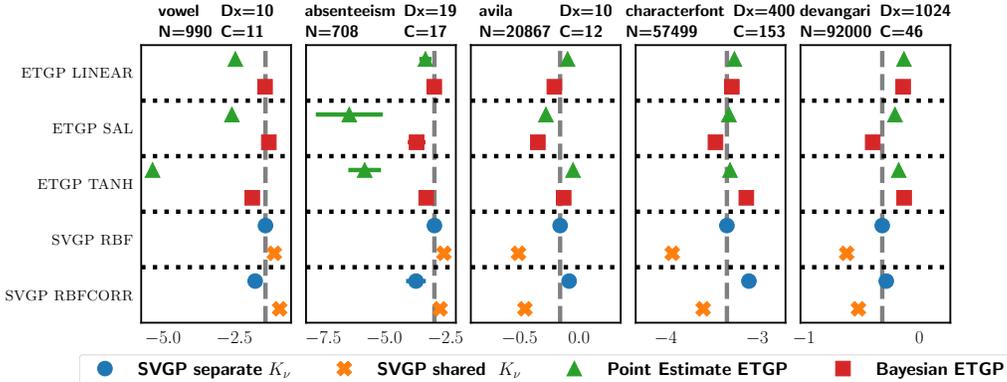}
    \caption{Log Likelihood (right is better) comparing the proposed model against independent/dependent stationary \GP{}s. }
    \label{fig:log-likelihood_stationary_mixing}
\end{figure}
\begin{figure}[!htb]
    \centering
    \includegraphics[width=\textwidth]{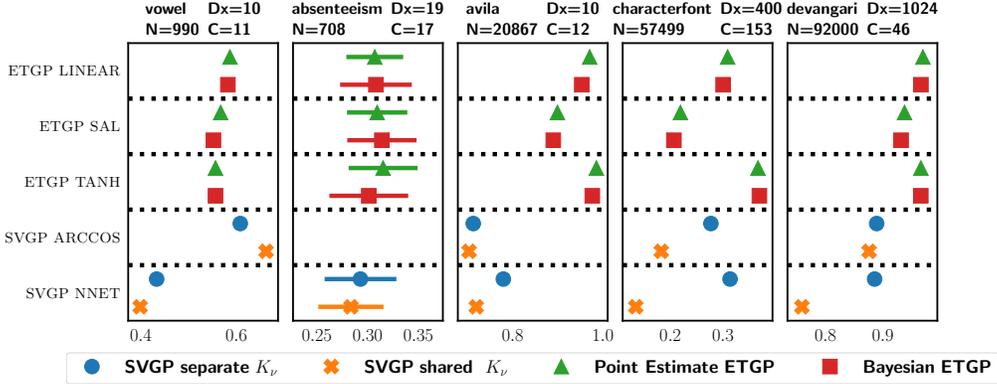}
     \caption{Log Likelihood (right is better) comparing the proposed model against non-stationary \GP{}s. }
    \label{fig:log-likelihood_non_stationary}
\end{figure}
\subsection{\SAL flow discussion}
In the work from \citep{TGP_maronas} input dependent \TGP{}s only used the \SAL flow parameterization. These flows have the good property that they can recover the identity function, see \citep{compositionally_warped_gp_rios:2019}, making them suitable for these applications since the marginal likelihood can penalize complexity in the warping function and `choose' to use a \GP by setting a linear mapping.

In this work, we have shown that \SAL flows provide, in general, worse results than the \TANH or \LINEAR flows. On one side, this shows the potential improvements that can be achieved by the versatility of the different flows that can be parameterized. Beyond being able to control moments of the induced distributions \citep{compositionally_warped_gp_rios:2019}, different flows combination can be more expressive or more training stable.

In particular, a problem with the \SAL flow is that small changes in its parameters can lead to mappings with a high derivative (like an exponential curve). This implies a big change in the gradients and thus while training we can suffer either from numerical saturation or convergence issues. This can be a possible explanation for this performance drop. In fact, in our experiments, we found \SAL to be the most saturating flow.

Nevertheless, exploring architectures with fewer parameters, such as \SAL flows, that can learn arbitrary non linear mappings such as those of \TANH, are an interesting line of research. For example the \TANH flow used has $12$ parameter, which implies an output \NN layer of $1836$ neurons for $C=153$ which considerably increases the computational burden when making Bayesian predictions, and also affects the speed improvement obtained. Anyway, this flow has similar training runs as \SVGP while providing much better metrics.
\subsection{Prediction time results}
We provide the average prediction time results in \fig \ref{fig:average_pred_time}, where we can see a similar trend as in the training time results. Our model can be one order of magnitude faster while providing similar or better prediction results. We can see how the reimplementation discussed in \uappendix \ref{gpflow_source_code_refactorization} boost the computational performance of the \SVGP \RBFCORR considerably.
\begin{figure}[!t]
    \centering
    \includegraphics[width=\textwidth]{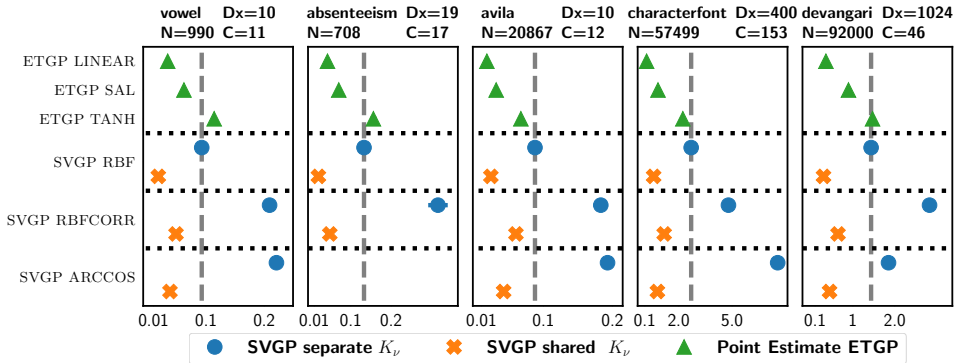}
     \caption{Average training time per epoch in minutes (left is better) comparing \ETGP{} with \SVGP{}s. \NNET kernel is omitted as it is slower.  }
    \label{fig:average_pred_time}
\end{figure}
\subsection{Using less inducing points}

In the final part of the appendix we provide additional results with less inducing points. We provide results with $50,\, 20$ inducing points and the \LL in \fig \ref{fig:less_inducing_points_complete_acc} and \fig \ref{fig:less_inducing_points_complete_ll}.

We observe how using less inducing points can regularize or provide an improvement over the \SVGP. We also see how the Bayesian flow provides, in the fewer inducing points setting, better results than the point estimate. This is specially significant in terms of \LL.

\begin{figure}[!t]
    \centering
     \includegraphics[width=\textwidth]{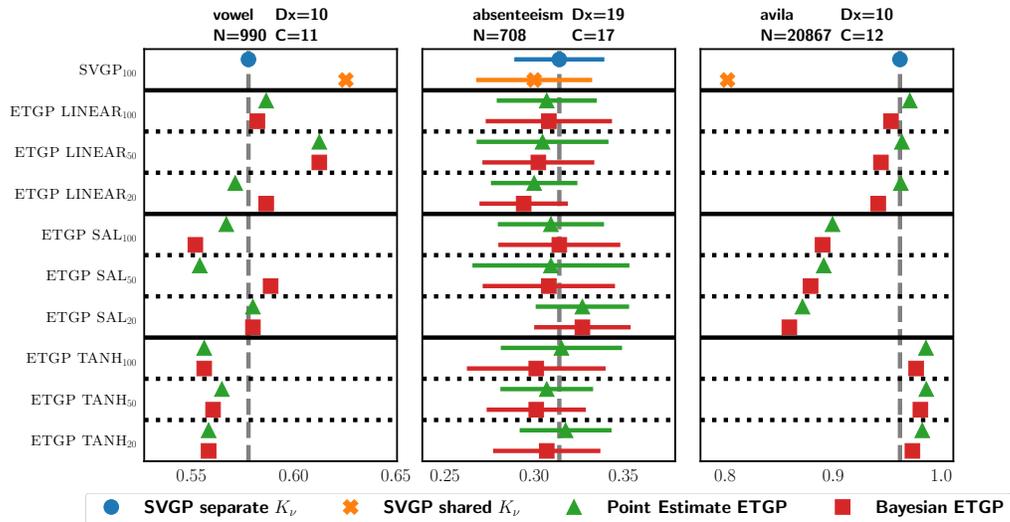}
    \caption{Results comparing \ACC (right is better) using less inducing points.}
    \label{fig:less_inducing_points_complete_acc}
\end{figure}

\begin{figure}[!t]
    \centering
     \includegraphics[width=\textwidth]{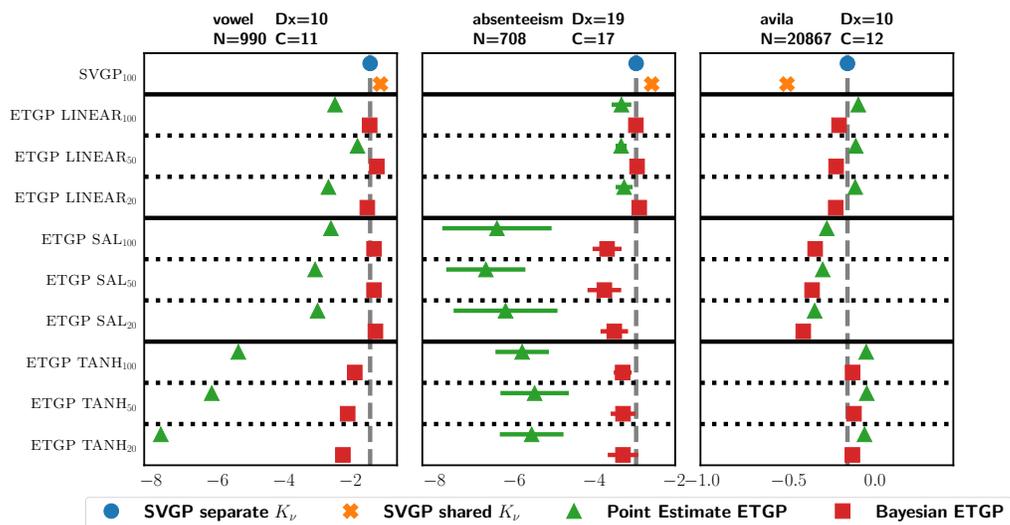}
    \caption{Results comparing \LL (right is better) using less inducing points.}
    \label{fig:less_inducing_points_complete_ll}
\end{figure}
}
\fi

%% file: sections/Z3_refactorizing_gpflow_source_code.tex
\section{Refactorizing \GPFLOW source code}
\label{gpflow_source_code_refactorization}

\ifapendixsubmission
{This information is attached in the supplementary  material.}
\else
{

In our experiments we found an issue in the source code of \GPFLOW, which considerably speed-up the performance of \SVGP{}s where the kernel is shared. We used \GPFLOW version $2.1.3$ and have found that the current issue also appears in the last stable version $2.5.2$. However, since the source code has considerably changed, our provided implementation is only valid for the earlier version.

We refactorize the source code in a way that is clear where the issue comes from, but note that better refactorizations could be done but would require a high-level and careful check of all the elements involved and how it would generalize to other computations.

The computation is concerned with the evaluation of the marginal variational distribution:
\begin{equation}
\begin{split}
     q(\fnbnull{c}) = \mathcal{N}( \fnbpos{0,n}{c} \mid & \GPcovar{c}{_{\X{},\Zsamples{c}}} \GPcovar{c}{_{\Zsamples{c},\Zsamples{c}}}^{-1}\varm{c} ,  \\
     &  \GPcovar{c}{_{\X{},\X{}}}-\GPcovar{c}{_{\Zsamples{c},\Zsamples{c}}}^{-1}[\GPcovar{c}{_{\Zsamples{c},\Zsamples{c}}}+\varS{c}]\GPcovar{c}{_{\Zsamples{c},\Zsamples{c}}}^{-1}\GPcovar{c}{_{\Zsamples{c},\X{}}} ) 
\end{split}\nonumber
\end{equation}
 of each of the $C$ latent processes when they are later mixed by a mixing matrix.  If we see, for a batch of samples $\Xsamples$ we need to evaluate $\GPcovar{c}{_{\X{},\X{}}}$ $C$ times if the kernel is not shared and only $1$ if the kernel is shared. We found \GPFLOW to always evaluate it $C$ times regardless of the type of kernel when a mixing matrix is later applied. In \GPFLOW version $2.1.3$ we can see that the LinearCoregionalization kernel does not differentiate if the latent \GP is shared or not, see \href{https://github.com/GPflow/GPflow/blob/v2.1.3/gpflow/kernels/multioutput/kernels.py}{https://github.com/GPflow/GPflow/blob/v2.1.3/gpflow/kernels/multioutput/kernels.py} line $174$. Thus, when computing the covariance, we can see in line $130$ here \href{https://github.com/GPflow/GPflow/blob/v2.1.3/gpflow/conditionals/multioutput/conditionals.py}{https://github.com/GPflow/GPflow/blob/v2.1.3/gpflow/conditionals/multioutput/conditionals.py} that the evaluation is performed $C$ times when it would be easier to evaluate it just once, and then tile $C$ times, which is the basic refactorization we perform. Also note that if the inducing points are not shared, but the kernel is, \GPFLOW would also perform this inefficient computation (see line $129$ where the kernel instance is repeated $C$ times). When the inducing points are shared and an independent kernel (no mixing matrix) is used, then \GPFLOW runs computations correctly (see lines $46$ and $92$).
 
 We observed similar issues in the newest \GPFLOW version $2.5.2$. We see that the LinearCoregionalization kernel in \href{https://github.com/GPflow/GPflow/blob/v2.5.2/gpflow/kernels/multioutput/kernels.py}{https://github.com/GPflow/GPflow/blob/v2.5.2/gpflow/kernels/multioutput/kernels.py} line $199$ do not make distinctions on whether the latent kernel is shared or not and this kernel is still inheriting from "Combination" class. Thus, the dispatcher in \href{https://github.com/GPflow/GPflow/blob/v2.5.2/gpflow/posteriors.py}{https://github.com/GPflow/GPflow/blob/v2.5.2/gpflow/posteriors.py} will not perform the computation starting in $778$ (note it checks if the kernel is SharedIndependent but LinearCoregionalization is type Combination) but that starting in $791$, which again evaluates the kernel $C$ times.
 
 We found several improvements that can be done to \GPFLOW source code and we will open corresponding issues and list them under our Github implementation of this work. We think that this particular one could be easily solved by using a \LMC kernel for latent shared kernel and a \LMC for latent separate independent kernel; or by explicitly checking if the elements in the underlying kernel list share references; or by, easily, using a list with just one element if the kernel is shared and $C$ elements if the kernel is not shared; or by wrapping around SharedIndependent or SeparateIndependent kernels base classes instead of using a list containing the kernels.

Another option, as employed by \GPYTORCH \citep{gpytorchref}, could be to evaluate the kernel directly using batched computations. We believe that the computational bottleneck comes from performing the kernel evaluation using a loop in python. This is because computing the squared distance in most kernels has a linear cost, so at most the corresponding added complexity should be $\complexity{Cd}$, with $d$ the dimensionality of the data. However, we found that without doing our improvement the code was much slower. Thus we further believe that computing the squared distance using batched computations rather than loops would also bridge the gap between the computational times of the \SVGP with a separate kernel and the \ETGP. Nevertheless our model is still competitive or even provides better results and faster since cubic computations do not scale linearly with $C$ which in a quick code snippet in \TENSORFLOW we found to be $1$ order of magnitude slower for $C=153$ and $M=100$ and $5$ times slower for $C=10$. The code snippet is:

\clearpage
\begin{minted}[mathescape,
               linenos,
               numbersep=5pt,
               gobble=2,
               frame=lines,
               framesep=2mm]{python}
    import tensorflow as tf
    import time
    import numpy as np

    C    = 10
    runs = 100

    X_ = np.random.randn(1,100,100)
    X  = X_@X_.transpose(0,2,1)
    X  = tf.constant(X)

    Y_ = np.random.randn(C,100,100)
    Y  = Y_@Y_.transpose(0,2,1)
    Y  = tf.constant(Y)

    acc = 0.0
    for i in range(runs):
        start = time.process_time(); 
        tf.linalg.cholesky(X);
        acc += time.process_time()-start
    print(acc/runs)
    
    acc = 0.0
    for i in range(runs):
        start = time.process_time();
        tf.linalg.cholesky(Y);
        acc += time.process_time()-start
    print(acc/runs)

 \end{minted}

}
\fi